%% file: main.tex
\documentclass[runningheads]{llncs}
\usepackage{graphicx}
\usepackage{tikz}
\usepackage{comment}
\usepackage{amsmath,amssymb}
\usepackage{color}
\usepackage{multirow}
\usepackage{epsfig}
\usepackage{tabularx}
\usepackage{booktabs}
\usepackage{dsfont}
\usepackage{pifont}
\usepackage[linesnumbered,ruled,vlined]{algorithm2e}
\usepackage{textcomp,booktabs}
\usepackage{colortbl}
\usepackage{caption}
\usepackage{subcaption}
\usepackage{wrapfig}
\usepackage[pagebackref=true,breaklinks=true,letterpaper=true,colorlinks,bookmarks=false]{hyperref}
\usepackage[accsupp]{axessibility}
\include{math_notation}
\newcommand{\deltadist}{Delta Distillation~}
\newcommand{\tinyspace}{\vspace{0.3mm}}

\definecolor{Gray}{gray}{0.95}
\definecolor{bl}{cmyk}{.3,0,0,0}
\makeatletter
\DeclareRobustCommand\onedot{\futurelet\@let@token\@onedot}
\def\@onedot{\ifx\@let@token.\else.\null\fi\xspace}
\def\eg{\emph{e.g}\onedot} 
\def\ie{\emph{i.e}\onedot}

\makeatother

\SetKwComment{Comment}{\color{green!50!black}// }{}

\newcommand{\assign}{\leftarrow}
\newcommand{\var}{\texttt}

\SetKwProg{Function}{function}{}{}
\newcommand{\opensupplement}{
    \setcounter{section}{0}
    \renewcommand\thesection{\Alph{section}}
}
\newcommand{\closesupplement}{
    \renewcommand\thesection{\arabic{section}}
}
\begin{document}
\pagestyle{headings}
\mainmatter
\title{Delta Distillation for Efficient Video Processing}
\titlerunning{Delta Distillation for Efficient Video Processing}
\author{
Amirhossein Habibian\inst{1}
\and Haitam Ben Yahia\inst{1}
\and Davide Abati\inst{1}\and\\
Efstratios Gavves\inst{2}
\and Fatih Porikli\inst{1}
}
\authorrunning{Habibian et al.}
\institute{
$^1$Qualcomm AI Research\footnote{Qualcomm AI Research is an initiative of Qualcomm Technologies, Inc.}
\qquad
$^2$University of Amsterdam
\email{\{ahabibia,hyahia,dabati,fporikli\}@qti.qualcomm.com}\\
\email{egavves@uva.nl}
}
\maketitle
\input{text/abstract}
\section{Introduction}
\label{sec:introduction}
\input{text/introduction}
\section{Related work}
\label{sec:related_work}
\input{text/related_work}
\section{Delta Distillation}
\label{sec:method}
\input{text/method}
\section{Experiments}
\label{sec:experiments}
\input{text/exp/experiments}
\section{Conclusion}
\label{sec:conclusion}
\input{text/conclusion}
\bibliographystyle{splncs04}
\bibliography{egbib}
\clearpage
\opensupplement
\begin{center}
    \large \textbf{Delta Distillation for Efficient Video Processing\\Supplementary material}
\end{center}
\input{supplementary_text/algorithms}
\input{supplementary_text/segmentation_exps}
\input{supplementary_text/teacher_student_specs}
\input{supplementary_text/limitations}
\closesupplement
\end{document}

%% file: math_notation.tex
\def\F{\mathcal{F}}

\DeclareMathOperator\E{\mathop{{}\mathbb{E}}}

\def\R{\mathbb{R}}

\def\x{\mathbf{x}}
\def\y{\mathbf{y}}
\def\z{\mathbf{z}}
\def\xt{\x_t}
\def\yt{\y_t}
\def\zt{\z_t}
\def\dzt{\Delta\zt}
\def\dxt{\Delta\xt}
\def\ztilde{\tilde{\mathbf{z}}}

\def\fl{f^l}

\def\thetal{\mathbf{\theta}_l}

\def\ftheta{f_{\theta}}

\def\gphi{g_{\phi}}

\def\Ldd{\mathcal{L}_{dd}}
\def\Lcost{\mathcal{L}_{s}}
\def\Ltask{\mathcal{L}_{t}}

\newcommand{\norm}[2]{\left\|#1\right\|_{#2}}

%% file: text/abstract.tex
\begin{abstract}
This paper aims to accelerate video stream processing, such as object detection and semantic segmentation, by leveraging the temporal redundancies that exist between video frames. Instead of propagating and warping features using motion alignment, such as optical flow, we propose a novel knowledge distillation schema coined as Delta Distillation.
In our proposal, the student learns the variations in the teacher's intermediate features over time. We demonstrate that these temporal variations can be effectively distilled due to the temporal redundancies within video frames.
During inference, both teacher and student cooperate for providing predictions: the former by providing initial representations extracted only on the key-frame, and the latter by iteratively estimating and applying deltas for the successive frames.
Moreover, we consider various design choices to learn optimal student architectures including an end-to-end learnable architecture search.
By extensive experiments on a wide range of architectures, including the most efficient ones, we demonstrate that delta distillation sets a new state of the art in terms of accuracy vs. efficiency trade-off for semantic segmentation and object detection in videos. Finally, we show that, as a by-product, delta distillation improves the temporal consistency of the teacher model.
\end{abstract}

%% file: text/introduction.tex
\input{figures/cover_figure}
The goal of this paper is to accelerate the processing of video streams, such as object detection and semantic segmentation. 
Despite the great progress in the development of efficient architectures~\cite{fasterseg,li2017PruningFF,zhang2016acceleratingvd,fanet,hrnet,efficientdet}, highly accurate models are still too expensive to process video frames in real-time.
This aspect hinders the deployment of accurate models on constrained settings,~\ie mobile devices.

To this end, recent works apply accurate yet expensive models only on a subset of frames, referred to as key-frames, and process the remaining ones using a lighter architecture~\cite{zhu17dff,accel,flow_guided,li2018low,liu2018mobile,memory_guided}. 
The representations from the light model are then aggregated with the key-frame representations from the expensive model in a recurrent structure: this step is typically performed at a deep layer, in order to leverage the representation power of the expensive model. 
Due to the misalignment between the current frame and key-frame, this strategy proves to be effective only when explicit motion alignment is carried out, typically by means of optical flow warping~\cite{zhu17dff,accel,flow_guided}.
For this reason, feature aggregation is a viable solution under the assumption that the overhead for extracting motion vectors is lower than in computations within the bypassed feature extraction.
Although this condition is easy to meet for expensive backbones such as ResNet-101, it has become less reasonable with the development of efficient models such as EfficientDet~\cite{efficientdet} or HRNet~\cite{hrnet}.

This paper introduces a novel approach to leverage the redundancies in a video to speed up the inference. 
Our proposal does not rely on explicit motion alignment and is applicable to any architecture, including the most efficient ones,~\ie EfficientDet-D0~\cite{efficientdet}, HRNet~\cite{hrnet} and the very recent DDRNet~\cite{ddrnet}. 
Our approach, coined as~\deltadist, is based on knowledge distillation~\cite{distillationsurvey,hinton}, a popular technique to accelerate an expensive \textit{teacher} network by distilling it into a lightweight \textit{student}.
Given an expensive teacher processing only key-frames, for every block of layers, we instantiate a cheap student counterpart, that is fed with a pair of frames and regresses their corresponding difference (delta) in teacher activations.
During training, the teacher provides target deltas, and the regression error of the student is minimized by an $\ell_2$ objective.
During inference, the teacher provides the representations for the key-frame, and the student iteratively updates them by adding predicted deltas in the following frames.
Due to the dense interplay between the teacher and student, happening at every block of the network, delta distillation effectively aggregates the features across frames without any explicit motion alignment.

Delta distillation has major differences to the common knowledge distillation setting, where the student learns to regress the teacher features as in Fig.~\ref{fig:cover_fig} (left). 
Instead of \textit{distilling the features}, delta distillation aims for \emph{distilling the temporal changes in the features} as illustrated in Fig.~\ref{fig:cover_fig} (right).
Intuitively, instead of learning the feature space embedding of their teacher, the delta distillation students learn the manifolds generated by transitions between samples, which we assume to be smooth in the case of correlated video frames.
We therefore hypothesize - and verify experimentally - that delta distillation, as compared to feature distillation, allows for learning much cheaper student functions for comparable performance.
Moreover, in contrast to the common knowledge distillation that relies solely on the student network to process all the test samples, delta distillation leverages both teacher and student during the inference. 
This trait enables delta distillation to enjoy having more parameters (coming from both models) without increasing the computational cost, as each test sample is processed by either the teacher or student network.

We summarize our contributions as follows: 
\emph{i)} We propose delta distillation, a novel approach to accelerate video inference without any explicit motion compensation involved. 
\emph{ii)} We elaborate various design choices to learn optimal student architectures including an end-to-end learnable architecture search.
\emph{iii)} We conduct extensive experiments on two different tasks and a wide range of models, including the most efficient architectures. 
Our analysis demonstrates that delta distillation sets a new state of the art in terms of accuracy vs. efficiency trade-off for semantic segmentation and object detection in videos. 
\emph{iv)} We show that, as a by-product, delta distillation improves the temporal consistency of the teacher model, even though it is not explicitly optimized to do so.

%% file: figures/cover_figure.tex
\begin{figure}[t]
\centering
\includegraphics[width=\columnwidth]{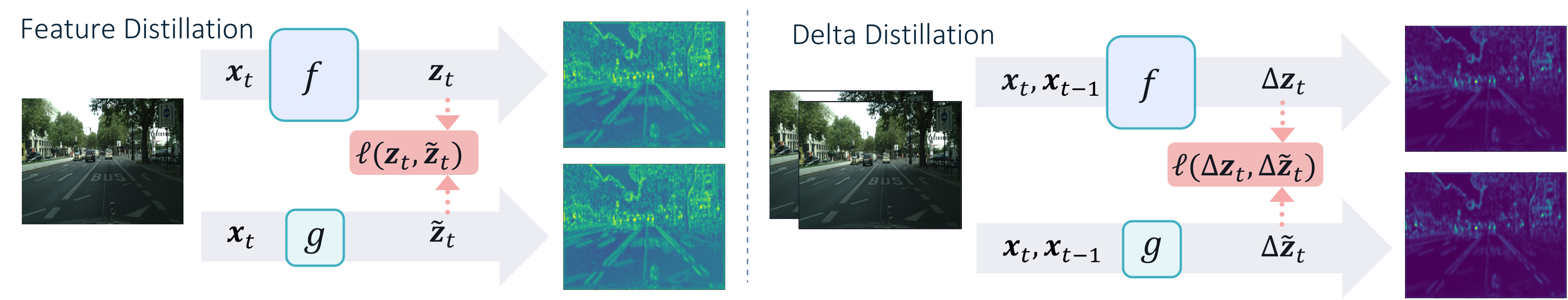}
\caption{Feature distillation vs. Delta distillation. Instead of distilling features $\z$ computed with an expensive layer $f$, we distill to a cheap student $g$ their changes across frames, $\Delta\z$. Due to temporal correlation in videos, transitions between frames are smooth and deltas are smaller (as visible) thus easier to distill.}
\vspace{-4mm}
\label{fig:cover_fig}
\end{figure}

%% file: text/related_work.tex
\subsubsection{Efficiency in deep learning}
Improving efficiency of neural networks is an active research area studied from multiple directions, comprising: quantization, to represent weights and activations with a low bit precision~\cite{quant2,quant3,krish2018quant,nagel2019dfq}, pruning, to discard unimportant or redundant channels~\cite{denil2013PredictingPI,comp2,li2017PruningFF}, neural architecture search, to find network designs with good accuracy vs. efficiency trade-offs~\cite{fasterseg,donna}, and low rank kernel decompositions~\cite{comp2,zhang2016acceleratingvd}.
However, for models operating on videos, redundancy among consecutive frames represents the most essential leverage to improve efficiency.
In recent years, several works investigated in this direction, and they represent the closest efforts to our proposal.
To avoid extracting expensive representations at every frame, feature aggregation using optical flow was explored~\cite{zhu17dff,accel,flow_guided}. 
Nevertheless, the application of such an approach on modern efficient architectures has become harder, as it requires careful model design and a proper cost balance between the feature extraction and motion alignment. 
Other works aim at building powerful representations over time, by aggregating features extracted by efficient models on past frames~\cite{li2018low,hu2020tdnet,liu2018mobile,memory_guided}.
These recurrent methods, however, prove more effective whenever explicit motion alignment operations are carried out~\cite{hu2020tdnet}, which incur an extra computational cost. 
Finally, sparse computation models limit the feature extraction to sparse spatial locations that change over time,~\eg at a pixel~\cite{skipconv} or at patch level~\cite{patchwork}.
However, this strategy is not robust to highly dynamic scenes, where most pixels change. Additionally, the theoretical compute gains do not always translate to latency improvement due to the inefficiency of sparse operations in most platforms.
\subsubsection{Knowledge distillation}
Another well established direction to accelerate deep neural networks is knowledge distillation~\cite{hinton}, where an efficient student network is optimized to match the output of an expansive teacher network or model ensemble.
This approach was then extended by performing such an optimization within network stages, effectively distilling intermediate functions rather than the output only~\cite{fitnets}.
After these seminal works, efforts have been spent towards online distillation methods, dropping the asynchronous training regimes of teacher and students in favor of a single optimization procedure.
For instance, in~\cite{zhang2018deep} multiple networks learn collaboratively without any teacher, and more recent works formalize the latter as ensemble of multiple students~\cite{zhu2018knowledge,guo2020online,wu2021peer}.
However, all these approaches do not specifically target video use-cases, and therefore transfer knowledge between models without explicitly distilling any temporal dynamic.
Differently, the proposed delta distillation directly operates on temporal changes of features and, as demonstrated by experiments, allows for much cheaper student models for comparable quality.

Furthermore, some recent works adapt the general framework to specific recognition tasks, for instance by selecting specific spatial locations for teacher-student distillation in for anchor-based object detection~\cite{dai2021general,fgfi}.
We hereby remark that our approach is task-agnostic and can be applied to any video task.

%% file: text/method.tex
We start with a teacher $\F$ as a \emph{spatial} network generating accurate representations for a given downstream task,~\eg HRNet~\cite{hrnet} or FasterRCNN~\cite{fasterrcnn}. Our goal is to distill this model into a more efficient \emph{spatio-temporal} equivalent. We first break $\F$ down into a composition of $L$ parametric blocks, as:
\begin{equation*}
\label{eq:sequential_architecture}
\F = f^L\circ \dots \circ f^2 \circ f^1.
\end{equation*}
We describe delta distillation for a single block, however highlighting that we carry out the procedure in all blocks within a given network.
The $l$-th block $\fl$ takes the form:
\begin{equation*}
\z^l = f^l_{\thetal}(\x^{l}),
\end{equation*}
where $\x^{l}$ and $\z^l$ denote the input and output of the block respectively, and $\thetal$ describes its learnable parameters. To simplify the notation, we will omit the block index $l$ as we focus on a single block ($\z = f_{\theta}(\x)$), and will reintroduce it in Sec.~\ref{sec:training}, where we define the overall network.
\subsubsection{Feature Distillation}
Feature distillation~\cite{fitnets} treats every block $\ftheta$ as a teacher block, providing target feature maps to supervise a student block $\gphi$, parametrized by $\phi$, typically designed to be much cheaper.
For instance, a distillation objective optimizes the expected $\ell_2$ norm of the error between $\ftheta$ and $\gphi$:
\begin{equation}
\label{eq:kd_objective}
\mathcal{L}_d(\x;\phi) = \E_{\x}\left[\norm{\ftheta(\x) - \gphi(\x)}{2}\right].
\end{equation}

\subsubsection{Delta Distillation}
Given a sequence of inputs $\xt$, the output of a block $\ftheta$ at a time-step $t$ can be written as:
\begin{equation*}
\label{eq:function_dz}
\zt = \z_{t-1} + \dzt,
\end{equation*}
where $\dzt$ represents how the output of the teacher changes over time. Considering the correlation between the consecutive samples in a video, we hypothesize that $\dzt$, being a transition function in the feature manifold, has a lower rank compared to the mapping $\ftheta$. For example, in the extreme case of identical frames, $\dzt$ will be of rank $0$.
We argue that $\dzt$ can be distilled more effectively than $\zt$ by a student with the same number of parameters, as verified by our experiments. In delta distillation, the student approximates the deltas given the current and previous frames as:
\begin{equation*}
\label{eq:student}
\Delta\ztilde_t \approx \gphi(\xt, \x_{t-1}).
\end{equation*}
This idea represents the core of our proposal that shifts the perspective from distilling the function, in Eq.~\ref{eq:kd_objective}, to distilling its temporal changes as:
\begin{equation}
\label{eq:kd_objective_delta}
\Ldd (\xt,\x_{t-1};\phi) = \E_{\xt,\zt}\left[\norm{\Delta \zt - \gphi(\xt, \x_{t-1})}{2}\right].
\end{equation}
By optimizing this objective, the student deltas $\Delta\ztilde_t$ converge to the teacher deltas $\dzt$ as shown in Fig.~\ref{fig:deltas}.\\
\input{figures/deltas}
\subsection{Student architectures}\label{sec:student_architectures}
As in any distillation approach, delta distillation admits diverse architectural choices~\cite{distillationsurvey}, in terms of \textit{i)} granularity at which it should operate, \eg distilling single convolutions vs distilling residual blocks or branches and \textit{ii)} possible architectures for the student block.
In our implementation, we consider the following:
\subsubsection{Linear blocks} that define the teachers and students at every convolution. In this case, we only feed the student with the input residual $\dxt = \xt - \x_{t-1}$:
\begin{equation*}
\gphi(\xt, \x_{t-1}) = \gphi^{conv}(\dxt).
\end{equation*}
This choice is motivated by the Taylor approximation of $\ftheta$ where, if the function $\ftheta$ is linear, only the first order term is non-zero, and the derivative $\nabla\ftheta(\x_{t-1})$ is a constant:
\begin{equation}
\label{eq:taylor}
\dzt = \nabla\ftheta(\x_{t-1})\dxt + \frac{1}{2}\nabla^2\ftheta(\x_{t-1})\dxt^2 + \dots
\end{equation}

As the student architecture, we rely on a spatial kernel factorization as depicted in Fig.~\ref{fig:svd_like}. Similar to the spatial SVD~\cite{spatial_svd}, we decompose each 2D kernel as a sequence of two 1D kernels while reducing the number of intermediate channels by a compression factor $\gamma$.
\input{figures/svd_like}

\subsubsection{Non-linear blocks} that define the teachers and students at a coarser granularity as a sequence of residual blocks. In this case, according to the Taylor approximation in Eq.~\ref{eq:taylor}, we parameterize the student as a function of both the previous input $\x_{t-1}$ and input residual $\dxt$ concatenated along the channel dimension:
\begin{equation*}
\gphi(\xt, \x_{t-1}) = \gphi^{block}(\x_{t-1},\dxt).
\end{equation*}
As the student architecture, we envision two strategies: \textit{i)} \textit{channel reduction}, where the student mirrors the teacher but with fewer channels: we add two pointwise convolution to the block as first and last layer to shrink and expand the channels respectively by a fixed factor. \textit{ii)} \textit{spatial reduction}, where the student resembles the teacher but operates on a smaller resolution: we add a strided pointwise convolution and a pixel shuffle up-sampling to the beginning and the end of each block, respectively.
\subsection{Student architecture search}
Different layers within a network may be compressible to different extents.
We empirically found that there might exist a few critical layers
\footnote{as an example, transition layers in HRNets~\cite{hrnet}} 
that are not amenable to distillation: if compressed, they hinder the model performance.
Therefore, instead of committing to the same student architecture for all the blocks, we introduce two candidates: \textit{i)} non-compressed architecture, identical to the teacher, preferred for hard-to-compress blocks. \textit{ii)} compressed architecture using the techniques introduced above for linear and non-linear blocks.

To find optimal student architectures, we introduce a learnable parameter $\psi\in\R^2$ per block, that is learned jointly with the student parameters. During training, architectures are sampled from a categorical distribution $q_{\psi}$ over the two candidate models, obtained by feeding $\psi$ to a softmax layer. We rely on the Gumbel-softmax~\cite{jang2016categorical,maddison2016concrete} reparametrization to estimate gradients.
To encourage the search algorithm to opt for the compressed architecture, we introduce a sparsity regularization objective as:
\begin{equation}
\label{eq:cost_loss}
\Lcost(;\psi) = \E_{\gphi\sim q_{\psi}} \left[\text{FLOPs}(\gphi)\right],
\end{equation}
where $\text{FLOPs}$ is the complexity of the sampled architecture $\gphi$ in terms of the number of floating point operations. This loss encourages the model to select the more efficient student architecture as much as possible.
If not regularized, $\psi$ would converge to the trivial solution of selecting the non-compressed architecture as it has a higher capacity and a better distillation performance.
\subsection{Training and Inference}
\label{sec:training}
\input{figures/inference}
Given a trained teacher network $\F$, comprising the teacher blocks, and a training set of labeled clips with $T$ frames $\{(\x_{1:T}, \y_{1:T})\}$, we train the delta distillation by optimizing the following objective:
\begin{equation}
\label{eq:overall_loss}
    \E_{\x_{1:T},\y_{1:T}} \left[\frac{1}{T}\sum_{t=1}^T \big(\Ltask +\alpha \sum_{l=1}^L \Ldd^l + \beta  \sum_{l=1}^L \Lcost^l\big)\right]
\end{equation}
where $\Ldd^l$ and $\Lcost^l$ denote the delta distillation and the sparsity regularization losses, computed for the $l$-th block, as defined in Eq.~\ref{eq:kd_objective_delta} and~\ref{eq:cost_loss} respectively. The hyper-parameters $\alpha$ and $\beta$ balance the contribution of the losses to learn a model yielding the best accuracy vs. efficiency trade-off.

$\Ltask(\xt, \yt;\Theta, \Phi)$ is the task loss used to train the network $\F$, where $\Theta=\{\theta_1 \dots \theta_L\}$ and $\Phi=\{\phi_1 \dots \phi_L\}$ denote all the learnable parameters in the teacher and student, respectively.
Although we envision an unsupervised training, by excluding $\Ltask$ from the objective, we conduct all our experiments while including $\Ltask$ as it yields a better performance.

As illustrated in Fig.~\ref{fig:dd_forward_pass}, delta distillation calls both teachers and students during the inference. More specifically, the key-frame is passed through the teacher to compute initial features. Then,  subsequent frames are fed to the students to update the previous features by predicting their deltas across the frames. This is different from typical knowledge distillation settings, such as feature distillation, where the teacher is only used during training and not during inference.
Pseudo code for training and inference is provided in the supplementary material.
\subsection{Temporal Consistency}
\label{sec:tc}
Another aspect of increasing importance for video streaming tasks is temporal consistency of model responses~\cite{lei2020blind,liu2020efficient,Rebol_2020_ACVRW}. Indeed, flickering predictions can make decision-making difficult in critical scenarios.
Although our main motivation is to improve the model efficiency, we argue that delta distillation can also improve the temporal consistency in predictions (as verified experimentally in Sec.~\ref{sec:segmentation_sota}).
As an explanation, we argue that delta distillation converts the \emph{spatial} teacher network to a \emph{spatio-temporal} model, as it propagates states from one timestep to the next, and explicit temporal dynamics improve the overall temporal stability, as also noted in several prior works~\cite{Rebol_2020_ACVRW,sibechi2019exploiting}.
Moreover, in delta distillation, the students have a regularization effect on the teacher.
More specifically, the teacher should generate smooth and low-rank features so as to be learnable by a student with a limited number of parameters. 
This penalizes the teacher from generating hard to distill deltas, \eg from representations that are inconsistent over time.

%% file: figures/deltas.tex
\bgroup
\setlength{\tabcolsep}{1.5pt}
\begin{figure}[t]
\centering
\resizebox{0.75\columnwidth}{!}{
\begin{tabular}{ccc}
\includegraphics[width=0.35\columnwidth]{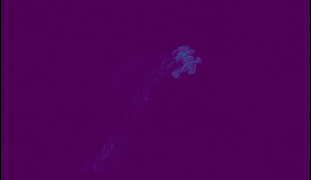}&
\includegraphics[width=0.35\columnwidth]{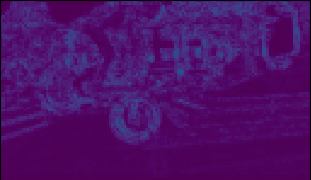}& \includegraphics[width=0.35\columnwidth]{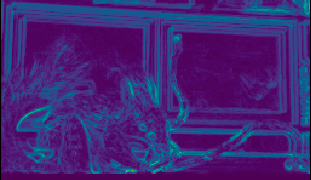}\\
\includegraphics[width=0.35\columnwidth]{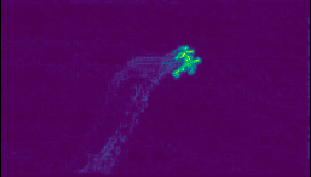}&
\includegraphics[width=0.35\columnwidth]{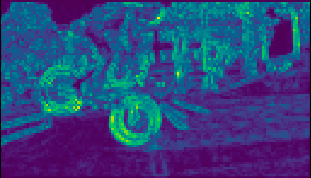}&
\includegraphics[width=0.35\columnwidth]{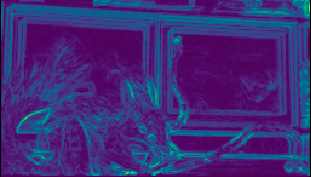}\\
Epoch 1 & Epoch 2 & Epoch 4
\end{tabular}}
\caption{The student deltas $\Delta\ztilde_t$ (top) vs. teacher deltas $\dzt$ (bottom). Thorough training the student learns to approximate the deltas from its teacher.}
\vspace{-4mm}
\label{fig:deltas}
\end{figure}
\egroup

%% file: figures/svd_like.tex
\begin{figure}[t!]
\centering
\includegraphics[width=0.45\columnwidth]{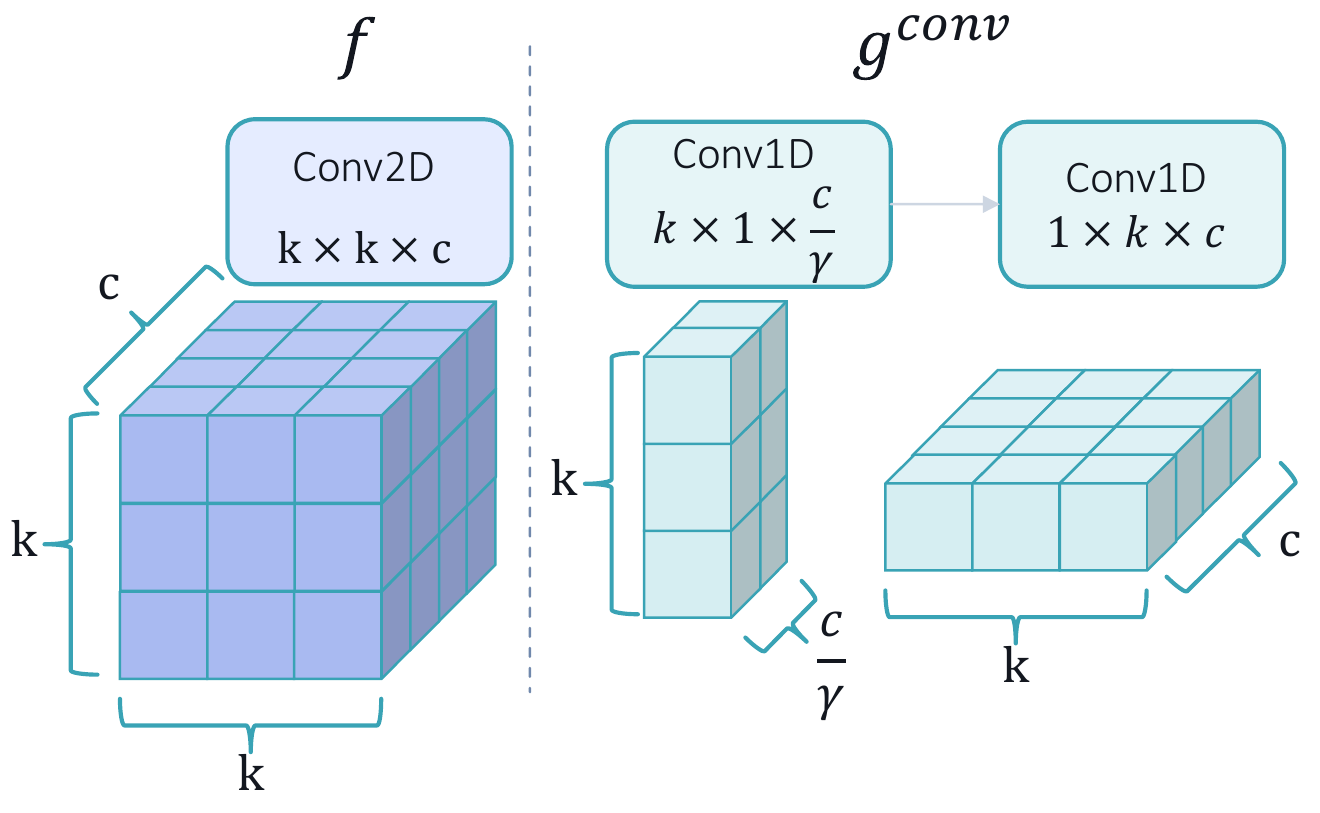}
\vspace{-2mm}
\caption{Student architecture for a linear block, obtained by decomposing the teacher kernel as a sequence of two 1D kernels with $\gamma \times$ less number of intermediate channels. For simplicity, we only visualize the output channels.
}
\vspace{-4mm}
\label{fig:svd_like}
\end{figure}

%% file: figures/inference.tex
\begin{figure}[t!]
\centering
\includegraphics[width=0.7\columnwidth]{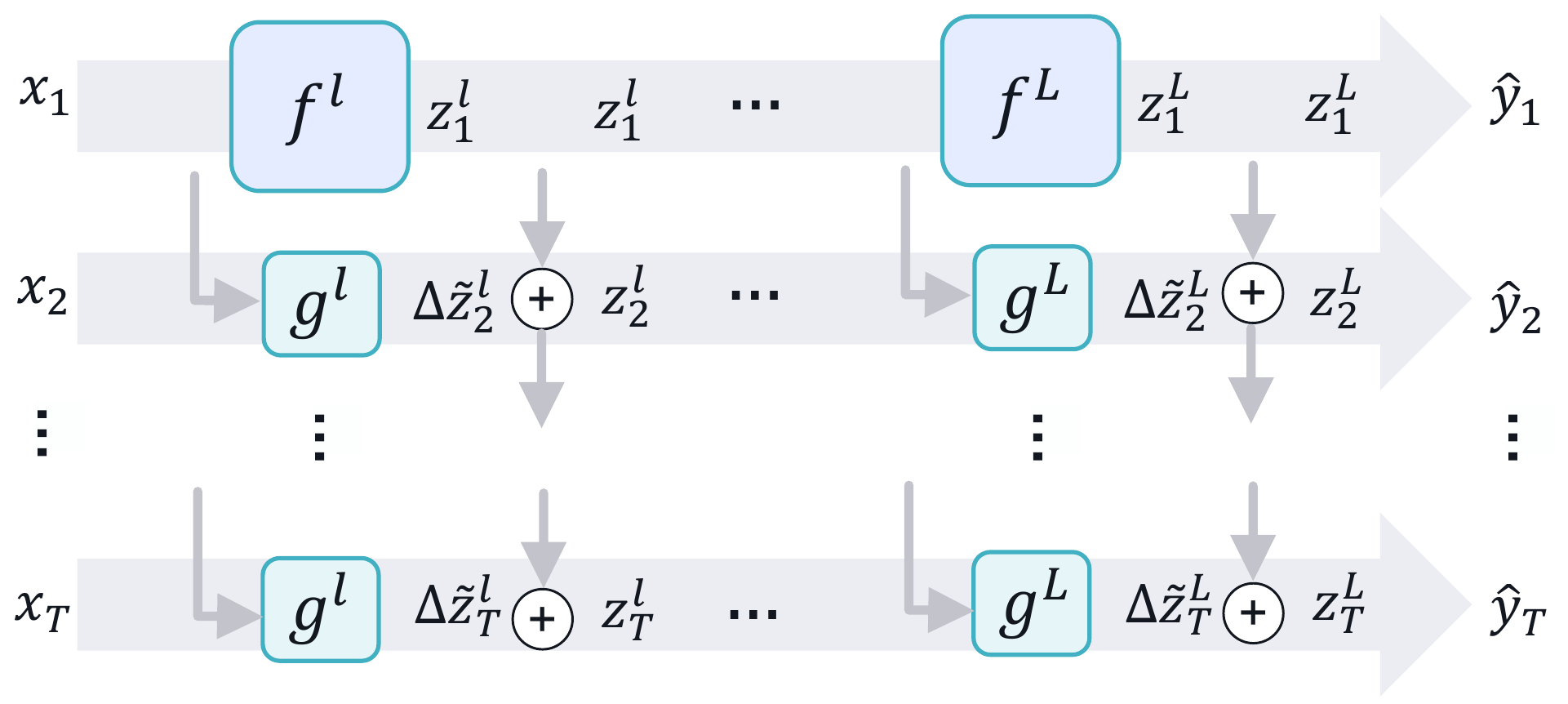}
\caption{Delta distillation at inference.
The teacher computes the features for the key-frame, while the student updates the features by predicting deltas over time.}
\label{fig:dd_forward_pass}
\vspace{-4mm}
\end{figure}

%% file: text/exp/experiments.tex
We evaluate delta distillation on two video tasks: object detection and semantic segmentation, in Sec.~\ref{sec:exp_detection} and ~\ref{sec:exp_segmentation}, respectively. Several ablation studies are reported in Sec.~\ref{sec:exp_ablation}. Further analysis are provided in the supplementary material.
\subsection{Object Detection}
\label{sec:exp_detection}
\input{text/exp/detection}
\subsection{Semantic Segmentation}
\label{sec:exp_segmentation}
\input{text/exp/segmentation}
\subsection{Ablation studies}
\label{sec:exp_ablation}
\input{text/exp/ablation}

%% file: text/exp/detection.tex
\subsubsection{Dataset and metrics}
We experiment with Imagenet VID~\cite{vid}, which contains 3862, 555, and 937 annotated snippets for training, validation, and test, respectively. All frames come with bounding boxes belonging to $30$ target categories.
Following the standard protocol~\cite{mega,flow_guided,fgfa,zhu17dff}, we augment the training snippets with still images from ImageNet DET sampled at $1:1$ ratio, and report results on the validation set. We rely on mean Average Precision (mAP) with an IoU threshold of $0.5$ as the accuracy metric.
To be hardware-agnostic, we report the efficiency of models in terms of number of floating point operations (FLOPs\footnote{FLOPs denotes number of multiply-adds.}).

\subsubsection{Training details}
We conduct our experiments using a single-stage and a two-stage object detector, namely EfficientDet-D0~\cite{efficientdet} and Faster-RCNN~\cite{fasterrcnn} with a ResNet-101 backbone.
We first train the teacher networks $\F$ using a SGD optimizer with a learning rate of $0.01$ for $7$ epochs. The learning rate is reduced by a factor of 10 at epochs $2$ and $5$. Clips are resized to have a longer side of $512~px$ and $600\sim1000~px$ for EfficientDet and Faster-RCNN, respectively. We train the models on four GPUs using a batch size of $16$ and $4$ for the EfficientDet and Faster-RCNN, respectively.
Starting from the trained teacher network and randomly initialized students, we optimize the training objective, Eq.~\ref{eq:overall_loss}, setting $\alpha$ and $\beta$ to $1000$ and $10$ for Faster-RCNN, and to $100$ and $10$ for EfficientDet. We rely on linear blocks with a compression ratio $\gamma=16$, as illustrated in Fig.~\ref{fig:svd_like}, to define the student architectures.
\input{figures/distillations_detection}
\subsubsection{Evaluation details}
Following~\cite{zhu17dff}, we split each video into sequences of equal length $T=10$. The first frame in a sequence (key-frame) is processed by the teacher while other frames are processed by the student networks. We report mAP and FLOPs averaged over all frames in the sequence.
In the supplement, we report detailed analysis teacher and student cost, illustrating per-frame FLOPS as well as their capacity in terms of parameters.
\subsubsection{Comparison to knowledge distillations}
We evaluate our key hypothesis, namely that $\dzt$ can be distilled more effectively than $\zt$, by training students using two different distillation objectives: feature vs. delta distillation as defined in Eq.~\ref{eq:kd_objective} and ~\ref{eq:kd_objective_delta}. Additionally, we compare to Fine Grained Feature Imitation (FGFI)~\cite{fgfi}, as a knowledge distillation devised for object detection that distills the features around the object anchor locations from a ResNet-101 backbone using a lighter student backbone (R50, R34, and R18). We omit applying FGFI on EfficientDet-D0 as there is no more efficient backbone available to serve as the student for this detector. We study how the model accuracy responds to reducing the computational cost~\ie by using a cheaper backbone for FGFI, and a higher sparsity regularization $\beta$ for feature and delta distillation.

As reported in Fig.~\ref{fig:feature_vs_delta}, our results demonstrate that FGFI is effective when compared to the optimization of the student backbones from scratch (blue plot), yet it underperforms with respect to the feature distillation that shares all components of our model, \eg SVD kernel decomposition and student architecture search. Additionally, the results verify that feature distillation has a limited effectiveness in reducing the model complexity especially for a highly optimized architecture such as EfficientDet-D0. However, by leveraging temporal redundancy, delta distillation reduces the compute cost by $2\times$ with a negligible drop in the accuracy. For EfficientDet-D0, it reduces the GFLOPs from 2.5 to 1.14 with a negligible mAP drop, from 69.0 to 68.8.
\subsubsection{Comparison to state of the art}
We compare to the state of the art video object detectors that leverage temporal redundancies to speed up the inference, as categorized into: \emph{i)} feature warping,~\ie DFF~\cite{zhu17dff} and Mobile-DFF~\cite{flow_guided}, that bypass feature computation by warping the previous features using optical flow.
\emph{ii)} feature aggregation,~\ie TAFM~\cite{liu2018mobile}, that instead of extracting expensive features at every frame, it relies on an aggregation of cheaper features over time.
\input{tables/sota_detection}
\emph{iii)} feature sparsification,~\ie PatchWork~\cite{patchwork} and Skip-Conv~\cite{skipconv}, that restrict the feature computation only to a sparse set of regions or pixels that change significantly across frames. \emph{vi)} detection by tracking,~\ie PatchNet~\cite{patchnet}, that interleaves running an expensive detector with cheap object trackers.
As reported in Tab.~\ref{tab:sota_detection}, we divide object detectors into two groups: light detectors (top) developed for mobile devices, using MobileNet-v2~\cite{mobilenetv2} and EfficientNet~\cite{tan2019efficientnet} backbones, and expensive detectors (bottom), using ResNet-101 backbone.

Our results show that delta distillation achieves the lowest FLOPs while being more accurate than both the light and expensive detectors.
Moreover, despite the differences between FasterRCNN-R101 and EfficientDet-D0 in terms of architectures and computational bottlenecks, delta distillation consistently reduces their FLOPs without any architectural adaption, ~\ie from 2.5 to 0.71 and from 149.1 to 29.2 respectively. This is not the case for feature warping methods, that require a careful architecture design to find a right cost and accuracy balance between feature extractor and optical flow model.

%% file: figures/distillations_detection.tex
\begin{figure}[t!]
\centering
\includegraphics[width=0.8\columnwidth]{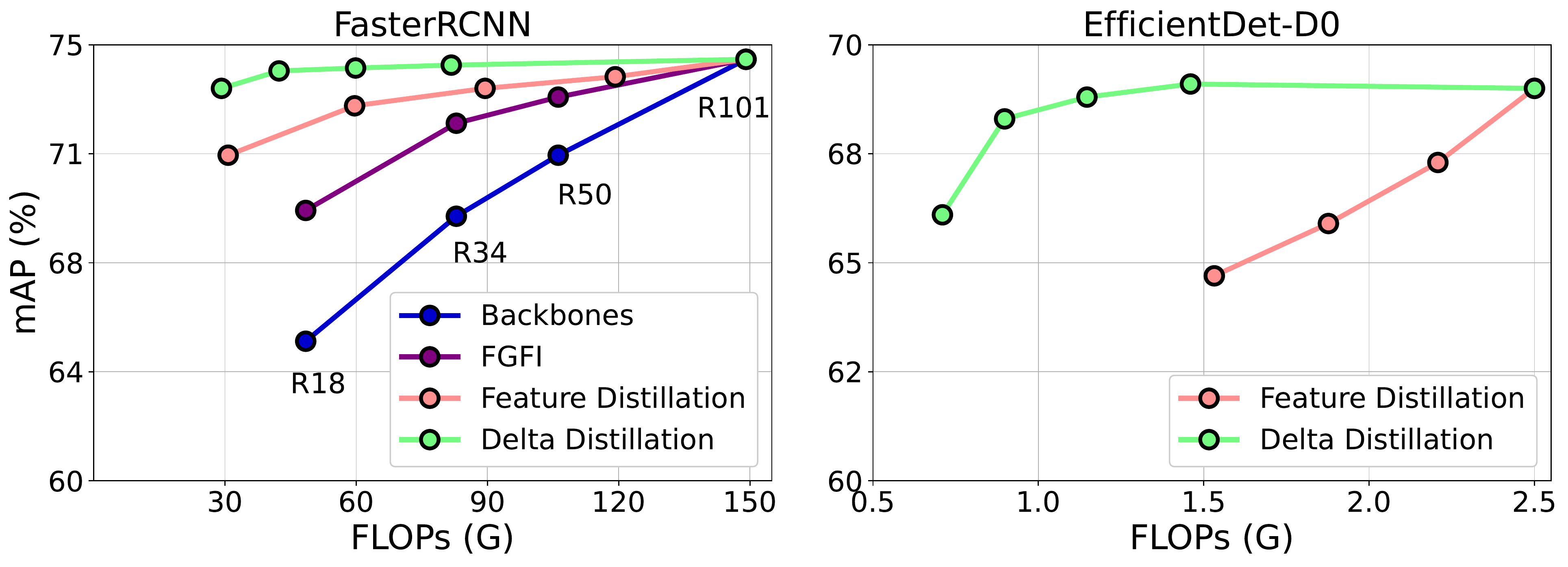}
\caption{Comparisons to knowledge distillations on ImageNet VID. Delta distillation outperforms the alternatives. The gap is higher for EfficientDet-D0 that is already highly optimized so is more challenging to be accelerated further.}
\label{fig:feature_vs_delta}
\vspace{-4mm}
\end{figure}

%% file: tables/sota_detection.tex
\renewcommand{\arraystretch}{1.1}
\begin{table}[tb]
\caption{Comparisons with efficient video object detectors on ImageNet VID, for some light (top) and heavy (bottom) networks. Delta distillation achieves the lowest FLOPs while being more accurate than others.}
\vspace{-0mm}
\begin{center}
\resizebox{0.45\columnwidth}{!}{
\begin{tabular}{lcc}
\toprule
\textbf{Model} &      \textbf{FLOPs (G)}     &  \textbf{mAP} \\ \hline
TAFM~\cite{liu2018mobile}        &   1.18    & 64.1\\
PatchWork~\cite{patchwork}       & 0.97 & 57.4 \\
Skip-Conv~\cite{skipconv}       &   0.78    & 62.9\\
PatchNet~\cite{patchnet}        &   0.73    & 58.9\\
Mobile-DFF~\cite{flow_guided}    & 0.71 & 62.8\\
\rowcolor{Gray}
\tinyspace
EfficientDet-D0~\cite{efficientdet}     &   2.50     & 69.0\\
\quad +~\textbf{\deltadist}              &  \textbf{ 0.71}    & 66.1\\
\midrule
DFF~\cite{zhu17dff}         &   34.9    & 72.5\\
PatchNet~\cite{patchnet}    &   34.2    & 73.1 \\
\rowcolor{Gray}
\tinyspace
Faster-RCNN~\cite{fasterrcnn}  &   149.1   & 74.5\\
\quad +~\textbf{\deltadist}         &   \textbf{29.2}    & 73.5\\
\bottomrule
\end{tabular}}
\end{center}
\label{tab:sota_detection} 
\vspace{-4mm}
\end{table}
\renewcommand{\arraystretch}{1.}

%% file: text/exp/segmentation.tex
\subsubsection{Dataset and metrics}
We conduct experiments on the Cityscapes dataset~\cite{cityscapes} that is partitioned into 2975, 500 and 1525 snippets as training, validation, and test splits respectively.
We rely on the standard training split to train and report results on the validation set.
The dataset provides pixel-level annotations into 19 classes for one frame per snippet.
We extract the remaining per-frame psuedo-annotations, which are required to train video models, by applying an off the shelf segmentation network~\cite{tao2020hierarchical} on unannotated frames in the training set.
We evaluate the accuracy using mean intersection-over-union (mIoU).
As a mean to compare computational cost, we rely again on FLOPs also reporting latency measurements in the supplementary materials.
\subsubsection{Training details}
\input{tables/sota_segmentation}
We conduct experiments using two state-of-the-art segmentation model: HRNet~\cite{hrnet} and DDRNet~\cite{ddrnet}.
We follow the same training protocol as in~\cite{ddrnet,hrnet}: models are initialized with ImageNet weights and trained using a SGD optimizer with a learning rate of 0.01, that is reduced using a polynomial decay policy with a power of 0.9. 
Training runs for 484 epochs using a batch size of 12 on four GPUs and SyncBN.
The models are trained on random crops of $512\times1024$ and tested on $1024\times2048$. For DDRNet~\cite{ddrnet}, we use online hard example mining~\cite{shrivastava2016training} as in~\cite{ddrnet}.
Starting from the trained teacher and randomly initialized students, we optimize Eq.~\ref{eq:overall_loss}, using $\alpha$ and $\beta$ set to 1 and 0.5 respectively.
We rely on linear blocks with a compression ratio $\gamma=4$, as illustrated in Fig.~\ref{fig:svd_like}, to define the student architectures.
\subsubsection{Evaluation details}
Each video is split into sequences of equal length $T$. We fix the sequence length $T=3$ as it yields the best trade-off between accuracy vs. efficiency though the model is relatively robust to longer sequences as reported in Fig~\ref{fig:temporal_robustness}.
Since the videos in Cityscapes have temporally sparse annotation, we repeat evaluations by opting each annotated frame in all possible positions within the sequence and report the averaged mIoU following~\cite{hu2020tdnet}. Similarly we report FLOPs averaged over all frames in the sequence counting for the both teacher and student costs.
\subsubsection{Comparison to state of the art}
\label{sec:segmentation_sota}
\input{figures/temporal_robustness}
We first assess the effect of delta distillation on different segmentation backbones with varying computational cost: HRNet-W18-small, DDRNet-23-slim, DDRNet-23, and DDRNet-39. 
As reported in Tab.~\ref{tab:sota_segmentation} (bottom), delta distillation consistently reduces computational cost by a factor of $\sim2\times$ for all the backbones, with no or small drop in accuracy.

Tab.~\ref{tab:sota_segmentation} compares delta distillation with efficient image\footnote{We limit our comparisons to efficient models with less than 100 GFLOPs.} (top) and video (middle) based semantic segmentation models.
The results show that delta distillation outperforms all the image-based models while being at the same time more efficient. 
Compared to BiseNet-v2~\cite{bisenetv2}, the most efficient frame-based model, delta distillation achieves a mIoU of 76.2 vs. 73.4, with a lower cost of 17.9 vs 21.1 GFLOPs. 
Moreover, delta distillation achieves a more favorable accuracy vs. efficiency trade-off compared to the other video-based models.
For instance, at the same mIoU of 79.9, delta distillation is $3.8\times$ more efficient (541 vs 140 GFLOPs) than a TDNet~\cite{hu2020tdnet} with PSPNet backbone. 
Finally, the delta distilled DDRNet-23-slim model significantly outperforms DFF~\cite{zhu17dff} and Skip-conv~\cite{skipconv}, both in terms of accuracy and efficiency.
Per-class analysis, reported in the supp. material, highlights that accuracies are retained on both static and dynamic classes.

\paragraph{Robustness to temporal variations.} We evaluate the impact of sequence length ($T$) on the performance of delta distillation as reported in Fig~\ref{fig:temporal_robustness}. The longer the sequence is, the less frequently features are refreshed by running the teacher model. 
This drops the accuracy as the student model has a limited capacity compared to the teacher. 
However, as highlighted in the results, the performance drop is smaller for delta distillation, especially on longer sequences, compared to competing methods. 
As a lower bound, we include an interleaved baseline that copies the predictions from the key-frame to consecutive frames. 
These findings verify the effectiveness of delta distillation in handling long range temporal variations.

\paragraph{Choice of backbone.} In Tab.~\ref{tab:sota_segmentation} we evaluate video-based models using the backbones originally used by authors~\ie ResNet-101 for DFF and ResNet-50 for TDNet-PSPNet. Since these backbones are too expensive to run in real-time we rely on more efficient backbones,~\eg DDRNet, that are arguably more challenging to be further accelerated. As a further analysis, we implement several video-based models using DDRNet-23-slim backbone as reported in Fig~\ref{fig:temporal_robustness}. The figure confirms the superiority of delta distillation as compared to alternative video efficiency models all using the same backbone.

\subsubsection{Temporal consistency}
\input{figures/tc_frames}
As motivated in Sec.~\ref{sec:tc}, we evaluate the effectiveness of delta distillation at improving the temporal consistency (TC) on the validation set. We follow the TC metric introduced in~\cite{liu2020efficient}: in a nutshell, it computes the average IoU among model predictions across successive frames, after motion compensation by optical flow warping.

Our results, presented in Tab.~\ref{tab:temp_consistency}, suggests several insights:
First, the training procedure of delta distillation effectively regularizes the teacher model towards more temporally consistent predictions, even when run on different frames independently (T): such a benefit is testified by the improvement of 1.8 points with respect to the baseline.
Furthermore, by using the inference procedure comprising of both teacher and student (T+S), the TC metric further improves. 
Finally, we compare to the temporal consistency reported by ETC~\cite{liu2020efficient} which explicitly includes a temporal consistency loss in the optimization.
Fig.~\ref{fig:tc_frames} shows qualitative results of DDRNet23 both with and without delta distillation.
\input{tables/tc_and_ablation_student}

%% file: tables/sota_segmentation.tex
\newlength{\oldintextsep}
\setlength{\oldintextsep}{\intextsep}
\setlength\intextsep{0pt}
\begin{wraptable}{r}{0.5\columnwidth}
\vspace{-4mm}
\begin{center}
\resizebox{0.5\columnwidth}{!}{ 
\begin{tabular}{lcc}
\toprule
\textbf{Model} &      \textbf{FLOPs (G)}     &  \textbf{mIoU} \\ \hline
FANet-34~\cite{fanet}               &65.0 & 76.3 \\
BiseNet-v1-18~\cite{bisenetv1}      &55.3 & 74.8 \\
FANet-18~\cite{fanet}               &49.0 & 75.0 \\
LedNet~\cite{wang2019lednet}        &45.8 & 71.5 \\
ICNet~\cite{icnet}                  &28.2 & 67.7 \\
FasterSeg~\cite{fasterseg}          &28.2 & 73.1 \\
ERFNet~\cite{romera2017erfnet}      &27.7 & 70.0 \\
SwiftNet-18~\cite{swiftnet}         &26.0 & 70.2 \\
BiseNet-v2~\cite{bisenetv2}         &21.1 & 73.4 \\
\midrule
TDNet-PSPNet~\cite{hu2020tdnet} &541.0 & 79.9 \\
DFF~\cite{zhu17dff}                 &109.3 & 69.2\\
TDNet-BiseNet~\cite{hu2020tdnet} & 101.3 & 76.4 \\
Skip-Conv~\cite{skipconv}          &29.0 & 75.5  \tinyspace\\
\midrule
\rowcolor{Gray}
DDRNet-39~\cite{ddrnet}         & 282.0 & 79.5 \\
\quad +~\textbf{\deltadist}     &140.0 & 79.9\tinyspace\\
\rowcolor{Gray}
DDRNet-23~\cite{ddrnet}         & 143.7 & 78.7 \\
\quad +~\textbf{\deltadist}     &71.8 & 78.9 \tinyspace\\
\rowcolor{Gray}
HRNet-w18-small~\cite{hrnet}    & 77.9 & 76.1 \\
\quad +~\textbf{\deltadist}     &34.1 & 75.7 \tinyspace\\
\rowcolor{Gray}
DDRNet-23-slim~\cite{ddrnet}    & 36.6 & 76.1 \\
\quad +~\textbf{\deltadist}     & \textbf{17.9} & 76.2 \\
\bottomrule
\end{tabular}}
\end{center}
\vspace{-4mm}
\caption{Comparison with efficient image (top) and video (middle) based models on Cityscapes validation set.}
\vspace{-4mm}
\label{tab:sota_segmentation} 
\end{wraptable}

%% file: figures/temporal_robustness.tex
\begin{figure}[t]
\centering
\includegraphics[width=0.45\columnwidth]{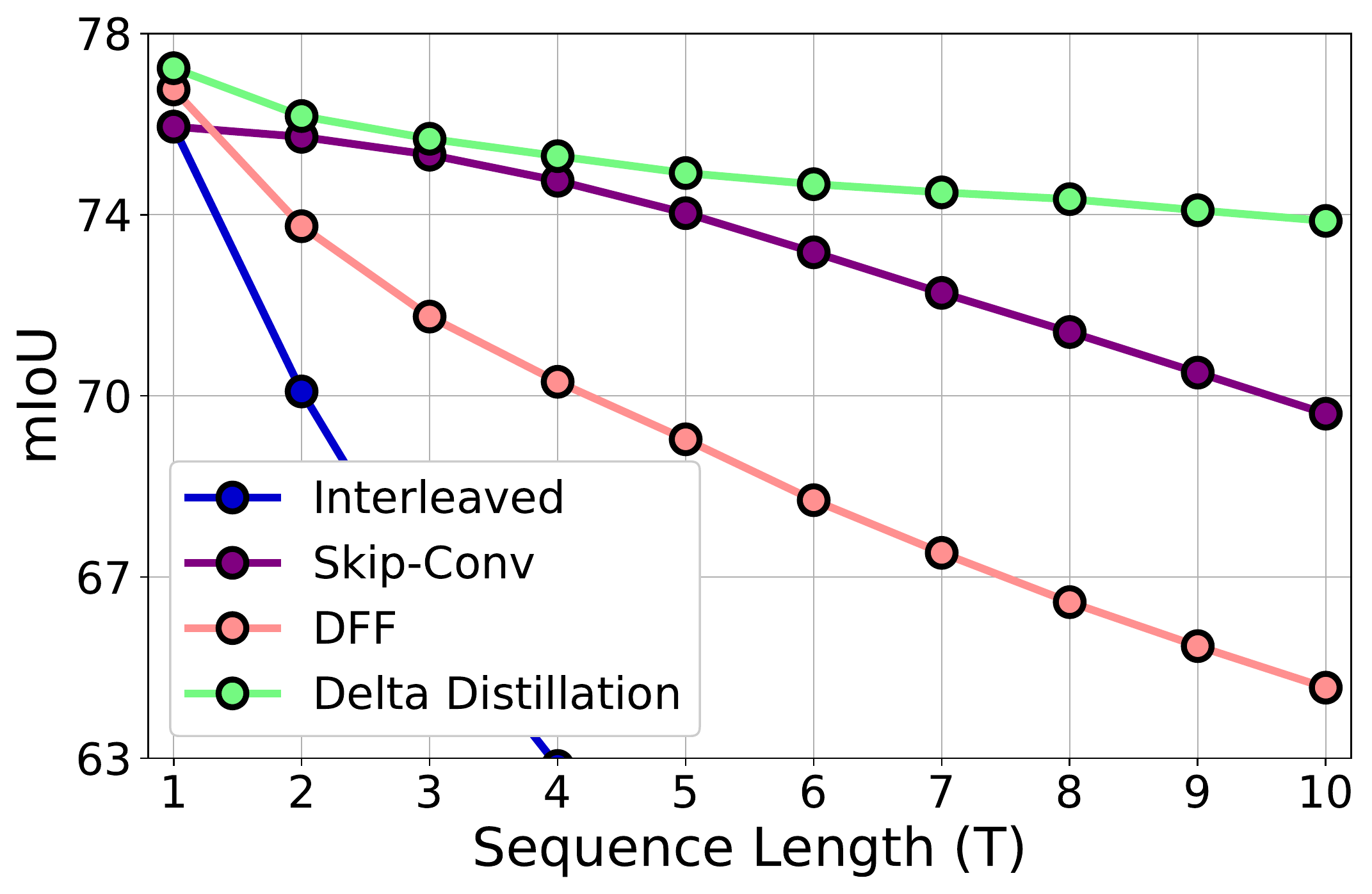}
\caption{Robustness to temporal variations. By increasing the distance to the key-frame, delta distillation retains better performances of prior methods.}
\label{fig:temporal_robustness}
\vspace{-6mm}
\end{figure}

%% file: figures/tc_frames.tex
\begin{figure}[t]
\centering
\includegraphics[width=\columnwidth]{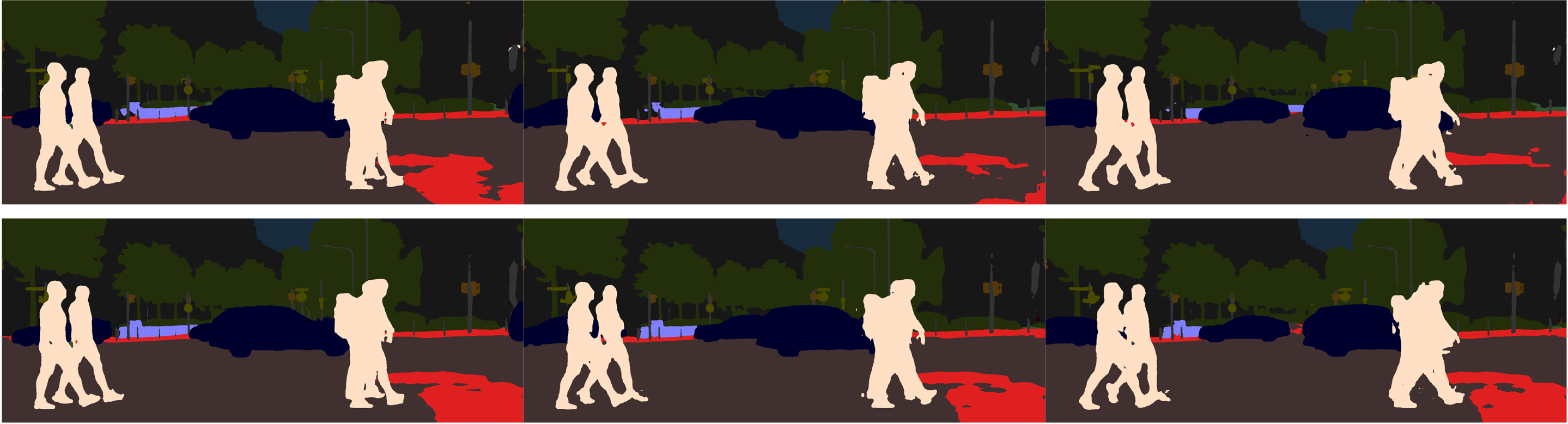}
\caption{Example predictions on Cityscapes, with DDRNet23 on top and DDRNet23 + \deltadist below. The latter shows more consistency over the baseline model.
More examples are reported in the supplementary material.}
\label{fig:tc_frames}
\vspace{-4mm}
\end{figure}

%% file: tables/tc_and_ablation_student.tex
\begin{table}[b!]
\begin{minipage}[t]{0.475\columnwidth}
    \vspace{-8mm}
    \caption{Temporal consistency (TC) measured on Cityscapes.}
    \centering
    \resizebox{0.8\columnwidth}{!}{
    \begin{tabular}{lcc}
    \toprule
    \textbf{Model}              & \textbf{TC} & \textbf{$\Delta$ TC} \\
    \midrule
    \rowcolor{Gray}
    PSPNET18~\cite{zhao2017pspnet}               & 83.3          &   -      \\
    \quad + ETC~\cite{liu2020efficient}          & 84.6          &  +1.3  \tinyspace\\
    \rowcolor{Gray}
    DDRNet23~\cite{ddrnet}                       & 82.6          &   -      \\
    \quad +~\textbf{\deltadist} (T)         & 84.4          &  +1.8    \\
    \quad +~\textbf{\deltadist} (T+S)         & \textbf{85.2} &  \textbf{+2.6}    \\
    \bottomrule
    \end{tabular}}
    \vspace{-4mm}
    \label{tab:temp_consistency}
\end{minipage}
\hfill
\begin{minipage}[t]{.475\columnwidth}
    \vspace{-8mm}
    \caption{Ablation on student architecture designs.}
    \centering
    \resizebox{1.\columnwidth}{!}{
    \begin{tabular}{lcc}
    \toprule
    \textbf{Student Architecture}              & \textbf{FLOPs (G)} & \textbf{mIoU}\\
    \midrule
    \rowcolor{Gray}
    DDRNet23             & 143.7       & 78.7\\
    \quad~+ Linear             & 71.8       & 78.9\\
    \quad~+ Non-Linear (Spatial)             & 110.3       & 78.4 \\
    \quad~+ Non-Linear (Channel)             & 84.3       & 79.0 \\
    \quad~+ Non-Linear (Channel + Spatial)  & 96.13       & 78.7 \\
    \bottomrule
    \end{tabular}}
    \vspace{-4mm}
    \label{tab:ablation_student_architecture} 
\end{minipage}
\end{table}

%% file: text/exp/ablation.tex
\subsubsection{Student architecture}
We analyze the effect of different choices of architecture, following the designs described in Sec.~\ref{sec:student_architectures}. 
In Tab.~\ref{tab:ablation_student_architecture}, we show the results of training \textit{Linear} vs \textit{Non-Linear} blocks as well as the \textit{channel} vs \textit{spatial} reduction.
First, we note that delta distillation, regardless of architectural choice, has lower GFLOPs than the base Image Model.
This trait is desirable, as it suggests the computational savings are not bound to a unique student architecture.
However, we do see some notable differences within architecture differences themselves.
By comparing both \textit{Linear} and \textit{Non-Linear - Channel}, we appreciate that the former enjoys a slightly smaller computational footprint (71.8 vs 84.3 GFLOPs) with similar mIoU. We hypothesize this difference could be due to the fact that linear functions are easier to distill.
Finally, when we compare the \textit{channel} and \textit{spatial} variants, we observe the latter architecture performs slightly worse.
\subsubsection{Student architecture search}
\input{figures/sparsity}
We study how the architecture search, discussed in Sec.~\ref{sec:student_architectures}, selects the student architectures. For this purpose, we observe the effect of gradually increasing the sparsity coefficient, $\beta$ from Eq.~\ref{eq:overall_loss}. We report the proportion of compressed blocks for DDRNet23-slim grouped by its four main stages: stem convolutions at the entry, low and high-resolution branches to process the input in parallel, and a pyramid pooling module (PPM) at the end to fuse feature maps across resolutions. We normalize the number of the compressed blocks per stage and report it as the compression rate in Fig.~\ref{fig:sparsity}.

We note that a higher $\beta$ indeed translates to a higher proportion of compressed blocks across all stages. Moreover, as we reduce $\beta$, the search algorithm opts for selecting less compressed blocks in the stem. We hypothesize that since all the layers follow the stem, this layer represents a single point of failure: at this stage, a non effective distillation might hinder the whole model's performance. We observe a similar pattern for the PPM stage, likely due to its closeness to the output, thus having a bigger impact on the performance.

%% file: figures/sparsity.tex
\begin{wrapfigure}{R}{0.5\columnwidth}
\centering
\includegraphics[width=0.5\columnwidth]{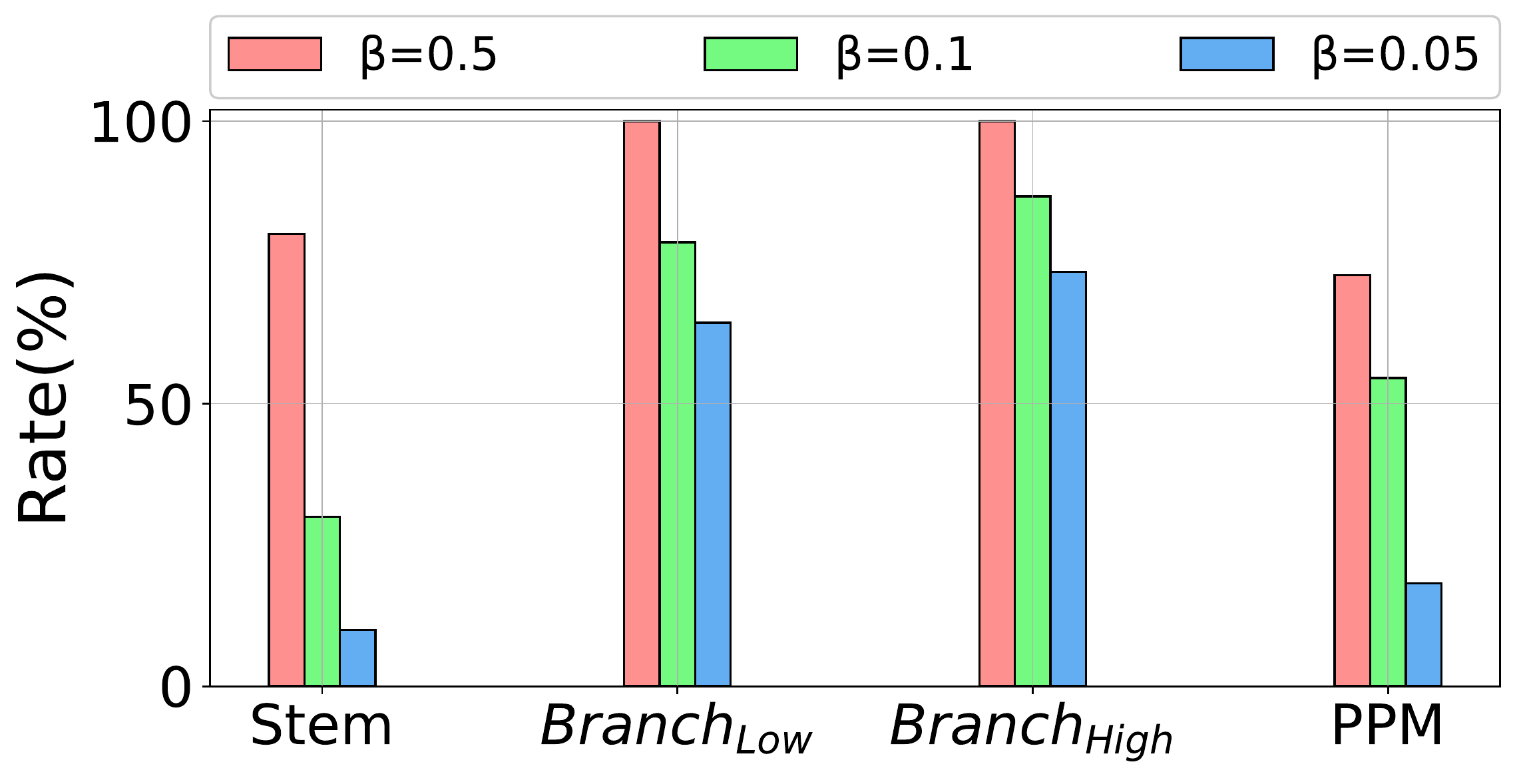}
\caption{Ablation on student architecture search.
We report the proportion of the compressed layers per stage.}
\label{fig:sparsity}
\end{wrapfigure}

%% file: text/conclusion.tex
We proposed delta distillation, a novel method for efficient video processing exploiting the temporal redundancy of frames.
Our proposal optimizes the regression, by means of cheap student models, of temporal variations in feature maps computed by an expensive teacher network.
During inference, the teacher provides the initial representations for the first frame; 
such feature maps are then iteratively refined for the next frames by adding deltas estimated by students, the latter operating at a low computational cost.
We show through extensive experiments that delta distillation outperforms feature distillation for comparable student architectures, and delivers state-of-the-art results for efficient video segmentation and object detection.
Moreover, we note how the proposed procedure improves the temporal consistency of the distilled model.

%% file: supplementary_text/algorithms.tex
\resizebox{\columnwidth}{!}{
\input{algorithms/train}
\input{algorithms/inference}
}
\section{Training and inference algorithms}
For clarity, we report the main pseudo-code algorithms characterizing delta distillation.
More specifically, Alg.~\ref{algo:training} illustrates the optimization of delta distillation objective $\Ldd$. 
Furthermore, Alg.~\ref{algo:inference} describes the inference schedule alternating between teacher (providing representations) and student (updating them).
We however recall that the full optimization of a delta distillation model, described in Sec.~3.3, also entails the optimization of the task objective $\Ltask$ and of the sparsity objective $\Lcost$, that we omit to avoid cluttering Alg.~\ref{algo:training}.

%% file: algorithms/train.tex
\begin{minipage}[t]{0.7\columnwidth}
\begin{algorithm}[H]
\small
\caption{Delta distillation (training $\Ldd$)}
\label{algo:training}
\Comment{$\eta$ is learning rate, globally given}
\Function{train\_dd($\ftheta$, $\gphi$, {\normalfont \var{x\_list}})}{
\Comment{compute groundtruth deltas}
\var{z\_list $\assign$ $\ftheta$(x\_list)}\;
\var{z\_cur $\assign$ z\_list[1:]}\;
\var{z\_past $\assign$ z\_list[:-1]}\;
\var{$\dzt$ = (z\_cur - z\_past).detach()}\;
\Comment{estimate deltas with student}
\var{x\_cur $\assign$ x\_list[1:]}\;
\var{x\_past $\assign$ x\_list[:-1]}\;
\var{$\Delta\ztilde_t = \gphi$(x\_cur, x\_past)}\;
\Comment{optimize distillation loss}
$\Ldd = \norm{\dzt - \Delta\ztilde_t}{2}$\;
$\phi \assign \phi - \eta\frac{\nabla\Ldd}{\nabla\phi}$\;
}
\end{algorithm}
\end{minipage}

%% file: algorithms/inference.tex
\begin{minipage}[t]{0.65\columnwidth}\vspace{0pt}
\begin{algorithm}[H]
\small
\caption{Delta distillation (inference)}\label{algo:inference}
\Comment{$T$ is globally given}
\Function{inference($\ftheta$, $\gphi$, {\normalfont \var{x\_list}})}{
\var{z\_list $\assign$ []}\;
\var{$\x^{t-1}, \z^{t-1} \assign $ none, none}\;
\For{$i\gets0$ \KwTo len(x\_list)}{
    \var{$\x^t \assign$ x\_list[i]}\;
    \If{$i$ $\%$ $T$ $==$ $0$}{
    \Comment{predict with teacher}
      $\z^t \assign \ftheta(\x^t)$\;
      }
    \Else{
        \Comment{update with student}
      $\z^t \assign \z^{t-1} + \gphi(\x^t, \x^{t-1})$\;
      }
    \var{z\_list.append($\z_t$)}\;
    $\x^{t-1}, \z^{t-1} \assign \x^t, \z^t$\;
    }
\Return{z\_list}
}
\end{algorithm}
\end{minipage}

%% file: supplementary_text/segmentation_exps.tex
\section{Further segmentation experiments}
\subsection{Latency measurements}
\input{tables/latency_comparison}
While the amount of GFLOPs is agnostic to hardware, software implementation, and measurement noise, it might not reflect latency gains exactly. 
In particular, it does not capture the memory usage overhead, which can become a limiting factor, especially for low-power devices.
To this end, we hereby report the latency of several image models for semantic segmentation, as well as the one from the corresponding delta distilled models.
More specifically, we report runtime measurements on simulated hardware of a mobile SoC, as this aligns well with the practical use cases of delta distillation. 
Indeed, resource constrained devices represent a real-world setting where research on efficient inference methods might contribute to significantly.
We illustrate such measurements in Tab.~\ref{tab:latency}.
In this experiment we opt, as a student architecture, for the non-linear channel reduction compression (see Sec.~3.1 in the main paper for details).
Specifically, for every architecture, we compress a sequence of two residual blocks in a row into a cheaper student, with a compression ratio $\gamma=4$.
Although delta distillation increases memory usage, due to transferring inputs and outputs of every block, we appreciate how the GFLOPS gains translate to latency improvements.
This behavior is more evident for heavier architectures (\ie DDRNet39 and DDRNet23), where the latency speedups are in line with (or even better than) the theoretical ones.
Improvements can still be appreciated when distilling cheaper architectures (\ie HRNet-w18-small and DDRNet23-slim), although they become less evident. 
\subsection{Per class IoU results}
\input{figures/per_class_iou}
We report in Fig.~\ref{fig:per_class_iou} the per-class IoU metric on the Cityscapes validation set, for three different variants of a DDRNet model, namely DDRNet23 slim (top), DDRNet23 (middle) and DDRNet39 (bottom).
The figure also reports performances of the corresponding delta distilled variant, with students instantiated at every linear block with a compression factor $\gamma=4$.
Interestingly, the figure testifies how the original and the compressed model behave similarily on all classes.
It is noteworthy how, for highly dynamic semantic classes such as cars, trains or bycicles, the student model can successfully update the representations from the teacher without any hurt in performance, showcasing that retained performances of the cheaper student model do not simply come from static classes.
We remark how this result is achieved without relying on any explicit motion compensation during training or inference.
\subsection{Temporal Consistency Results}
\input{figures/tc_frames_supplemental}
In Fig.~\ref{fig:tc_supplemental}, we illustrate further examples of the improved temporal consistency empowered by Delta Distillation. 
The figure, which serves as an extension of Fig.~7 in the main paper, showcases some segmentation results on the Cityscapes dataset, and it compares predictions from the original image-based architecture (DDRNet23) with the corresponding delta-distilled model.
As it can be appreciated, especially by comparing predictions within the marked dashed line boxes, Delta Distillation enjoys more temporal consistency in outputs, in regions where per-frame predictions result flickery.
We remark that temporal consistency is not an explicit objective of Delta Distillation, but rather a beneficial side effect enabled by its inference scheme, that relies on the update of past representations, in a procedure that is inherently more temporally smooth with respect to independent processing of frames.

%% file: tables/latency_comparison.tex
\begin{table}[t]
\caption{The efficiency gains by \deltadist translate to latency improvement in resource constrained devices.}
\begin{center}
\begin{tabular}{lm{.5cm}m{1cm}cm{.5cm}m{1cm}c}
\toprule
\multirow{2}{*}{\textbf{Model}} && \multicolumn{2}{c}{\textbf{FLOPs}} && \multicolumn{2}{c}{\textbf{Latency}}\\
&& \textbf{(G)} & $\uparrow$ && \textbf{(ms)} & $\uparrow$\\
\midrule
\rowcolor{Gray}
DDRNet39 && 282.0 & - && 41.0 & -\\
\quad+~\textbf{\deltadist} && 131.6 & (2.1$\times$) && 16.5 & (2.5$\times$)\\ 
\rowcolor{Gray}
DDRNet23 && 143.7 & - && 22.6 & -\\
\quad+~\textbf{\deltadist} && 71.3 & (2.0$\times$) && 12.4 & (1.8$\times$)\\ 
\rowcolor{Gray}
HRNet-w18-small && 77.9 & - && 39.4 & -\\
\quad+~\textbf{\deltadist} && 35.7 & (2.2$\times$) && 33.6 & (1.2$\times$)\\ 
\rowcolor{Gray}
DDRNet23-slim && 36.6 & - && 6.6 & -\\
\quad+~\textbf{\deltadist} && 13.7 & (2.7$\times$) && 5.1 & (1.3$\times$)\\ 
\bottomrule
\end{tabular}
\end{center}
\label{tab:latency} 
\end{table}

%% file: figures/per_class_iou.tex
\begin{figure}
    \centering
    \includegraphics[width=0.9\columnwidth]{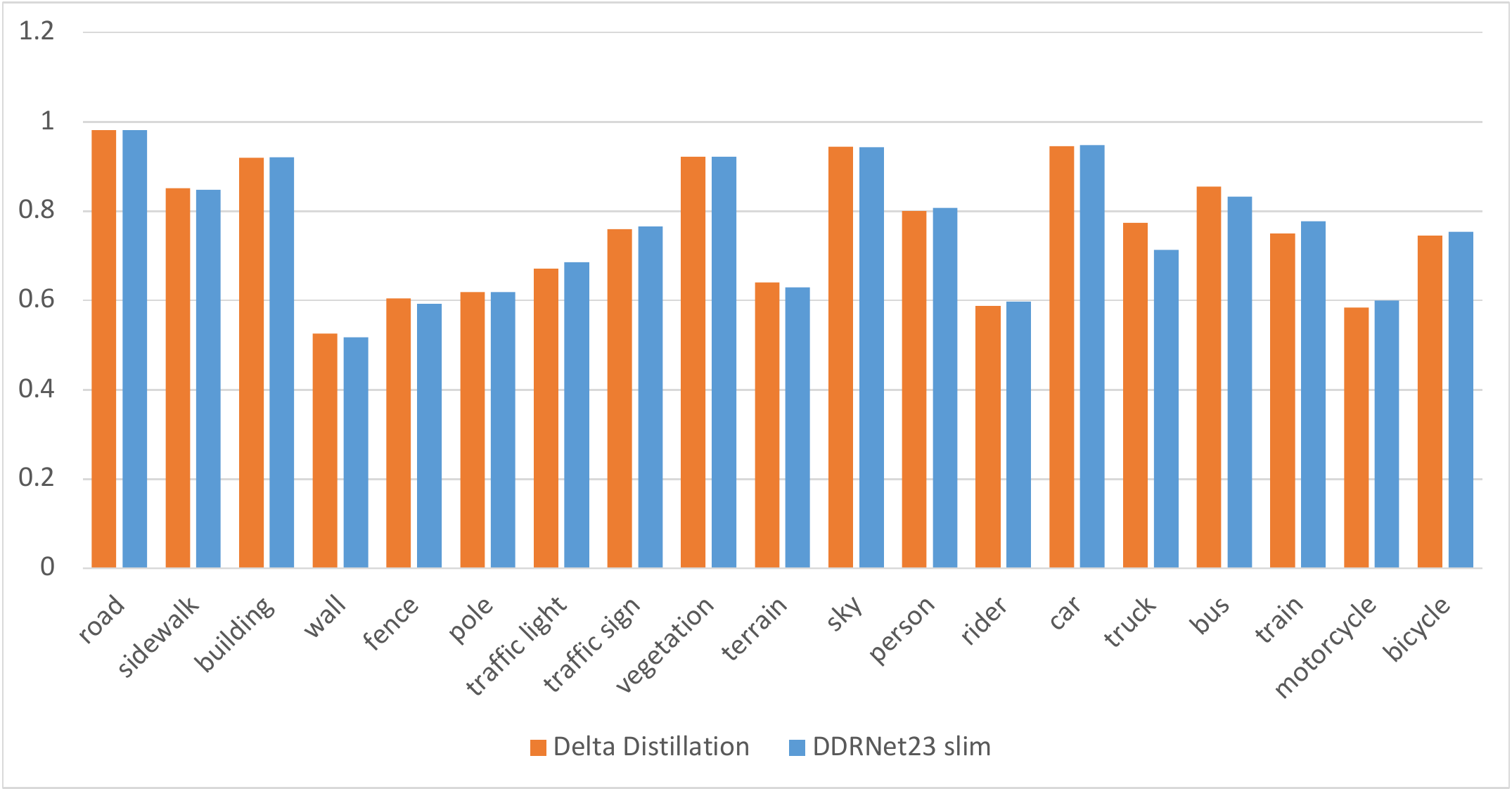}
    \includegraphics[width=0.9\columnwidth]{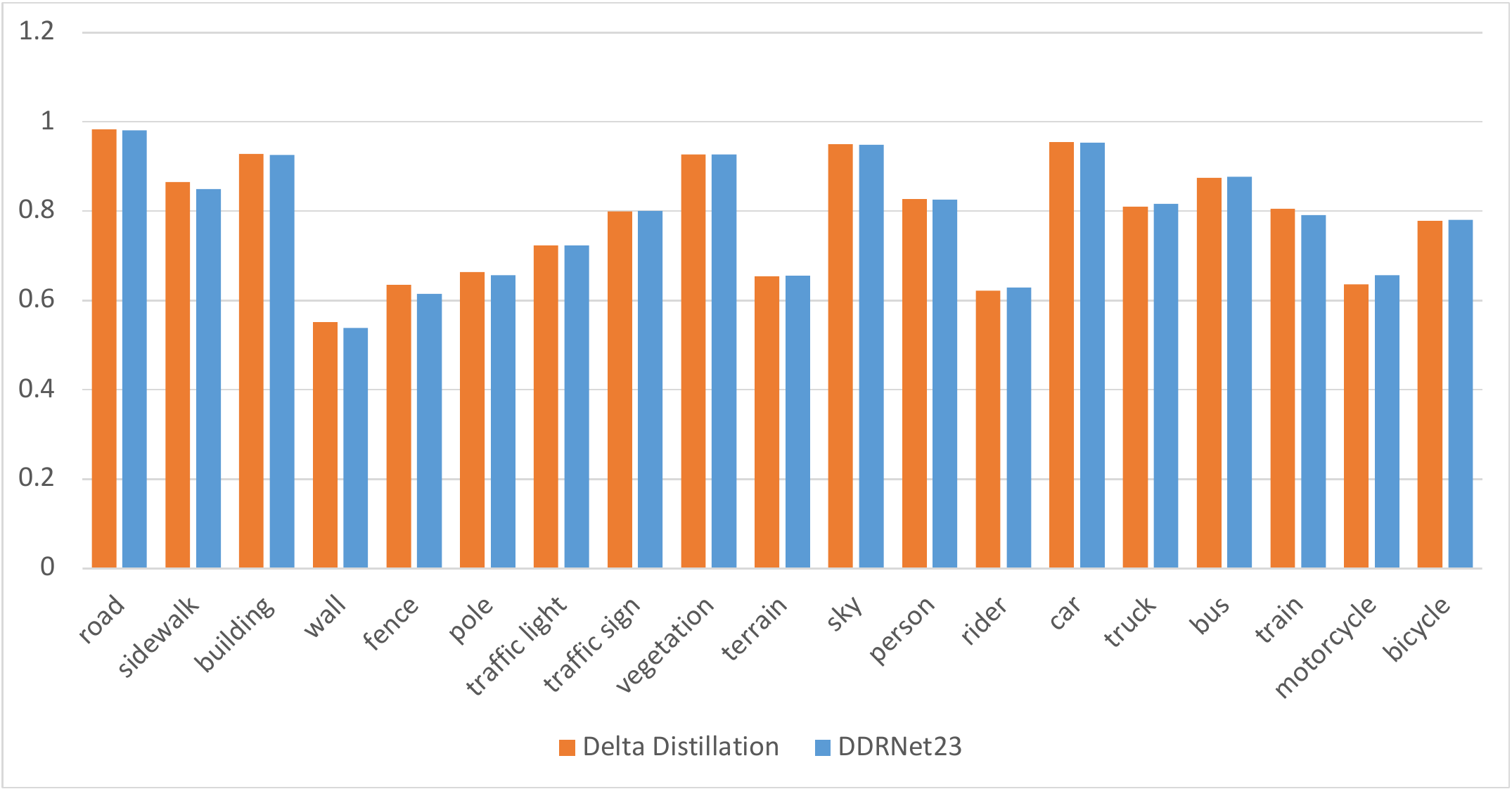}
    \includegraphics[width=0.9\columnwidth]{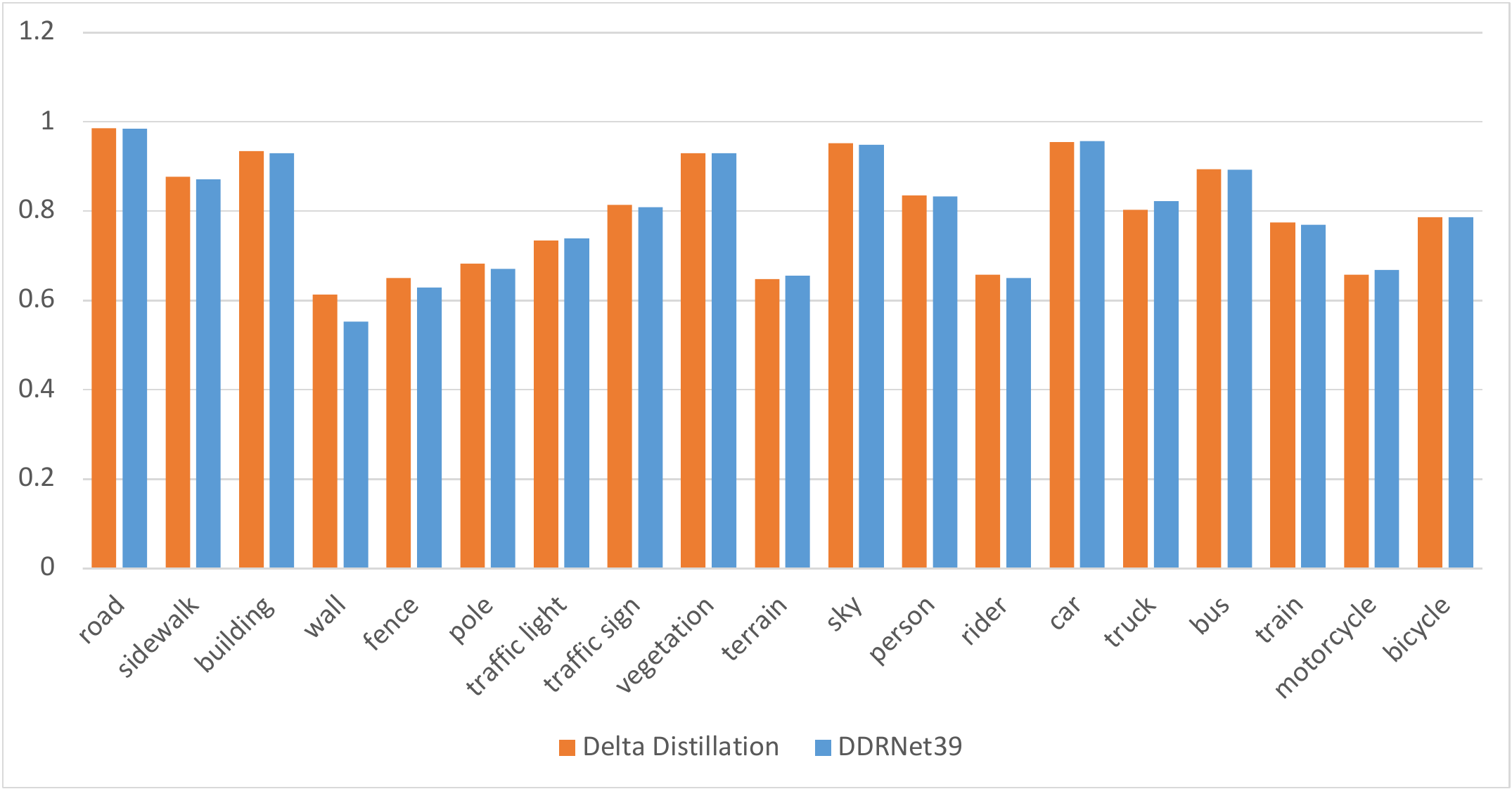}
    \caption{Per class IoU on Cityscapes for different DDRNet models. Note that Delta distillation does not exploit any single class, but rather improves upon several or retains overall IoU at a lower cost.}
    \label{fig:per_class_iou}
\end{figure}

%% file: figures/tc_frames_supplemental.tex
\begin{figure}[tbh]
    \centering
    \includegraphics[width=\columnwidth]{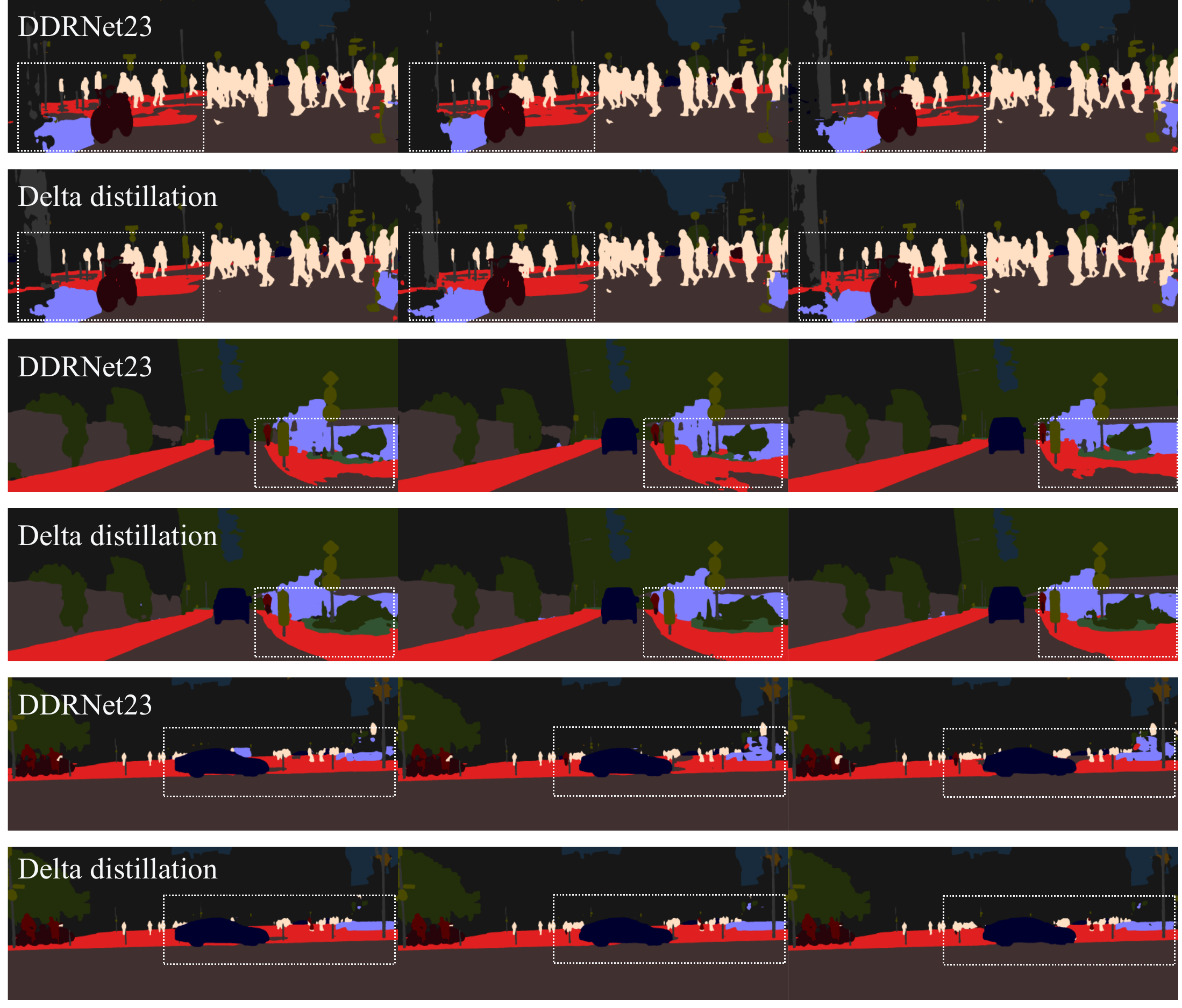}
    \caption{Additional results of DDRNet23 and Delta Distillation in alternating rows respectively. We highlight large differences between methods in the boxes with white dashed lines. We observe that delta distillation tries to maintain large areas assigned to particular classes over time more so than the baseline model.}
    \label{fig:tc_supplemental}
\end{figure}

%% file: supplementary_text/teacher_student_specs.tex
\section{Teacher and Student Parameters}
\input{tables/teacher_student_specs}
In all experiments in Sec.~4 in the main paper we report amortized GMACs as computational cost measure.
Therefore, the reported numbers illustrate the \emph{average} cost required to perform a prediction and, due to the schedule we apply at inference time assigning frames unevenly to different models, they are affected by \emph{i)} the teacher cost, \emph{ii)} the student cost and \emph{iii)} the schedule period $T$.
For clarity and completeness, we report in Tab.~\ref{tab:teacher_student_specs} the unamortized costs of the teacher and student for the main models used in the paper.
Additionally, the table reports the corresponding number of learnable parameters.

%% file: tables/teacher_student_specs.tex
\begin{table}[tb]
\centering
\caption{Specifications of the architectures used in Sec.~4 the main paper. We hereby report the Flops and the number of parameters separately for the teacher and student models.}
\label{tab:teacher_student_specs}
\begin{tabular}{cm{1em}llm{1em}rrc}
\toprule
\multirow{2}{*}{\textbf{Task}} && \multicolumn{2}{c}{\multirow{2}{*}{\textbf{Model}}} && \textbf{Flops} & \textbf{Params} & \textbf{Ref}\\
&& & && \textbf{(G)} & \textbf{(M)} & \textbf{(paper)}\\
\midrule
\multirow{3}{*}{Obj. Detection} && \multirow{3}{*}{EfficientDet-D0} & Teacher && 2.5 & 3.84 & \multirow{3}{*}{Tab.~1}\\
&& & Student && 0.5 & 1.12\\
&& & Amortized ($T$=10)&& 0.7 & -\\
\midrule
\multirow{3}{*}{Obj. Detection} && \multirow{3}{*}{Faster-RCNN} & Teacher && 149.1 & 81.22 & \multirow{3}{*}{Tab.~1}\\
&& & Student && 15.9 & 29.97\\
&& & Amortized ($T$=10) && 29.2 & -\\
\midrule
\multirow{3}{*}{Segmentation} && \multirow{3}{*}{DDRNet-39} & Teacher && 282.0 & 34.14 & \multirow{3}{*}{Tab.~2}\\
&& & Student && 68.9 & 8.09\\
&& & Amortized ($T$=3) && 140.0 & -\\
\midrule
\multirow{3}{*}{Segmentation} && \multirow{3}{*}{DDRNet-23} & Teacher && 143.7 & 20.79 & \multirow{3}{*}{Tab.~2}\\
&& & Student && 35.8 & 5.12\\
&& & Amortized ($T$=3) && 71.8 & -\\
\midrule
\multirow{3}{*}{Segmentation} && \multirow{3}{*}{DDRNet-23 slim} & Teacher && 36.6 & 5.91 & \multirow{3}{*}{Tab.~2}\\
&& & Student && 8.5 & 1.47\\
&& & Amortized ($T$=3) && 17.9 & -\\
\bottomrule
\end{tabular}
\end{table}

%% file: supplementary_text/limitations.tex
\section{Future work}
We recognize two limitations of our work, that we plan to tackle in future work.
First, delta distillation incurs some memory overhead: similarly to recurrent models, feature maps need to be stored in order to be propagated to the following time-steps.
The corresponding impact on memory thus depends on the granularity at which our model operates: it might be negligible when a few (bigger) blocks are optimized, whereas it could become meaningful when the distillation involves every layer of the original model.
Although we did not observe noticeable effects on latency (see Tab.~\ref{tab:latency}) in our experiments, such an overhead might be impactful for large models with many layers.

Moreover, our proposal is capable of reducing the amortized runtime of the original network, yet it features unbalanced latencies between frames.
Indeed, the inference schedule assigns frames either to the teacher or student model: frames processed by the former do not enjoy any latency reduction.
We envision this could be addressed by deploying the two models in separate threads, and by implementing an asynchronous inference scheme similar to~\cite{liu2019looking}.

In conclusion, we acknowledge our proposal introduces a few additional hyper-parameters into the network training schedule (\emph{i.e.}~$\alpha$, $\beta$), whose determination required some trial-and-error based tinkering.

%% file: main.bbl
\begin{thebibliography}{10}
\providecommand{\url}[1]{\texttt{#1}}
\providecommand{\urlprefix}{URL }
\providecommand{\doi}[1]{https://doi.org/#1}

\bibitem{patchwork}
Chai, Y.: Patchwork: A patch-wise attention network for efficient object
  detection and segmentation in video streams. In: ICCV (2019)

\bibitem{fasterseg}
Chen, W., Gong, X., Liu, X., Zhang, Q., Li, Y., Wang, Z.: Fasterseg: Searching
  for faster real-time semantic segmentation. ICLR  (2020)

\bibitem{mega}
Chen, Y., Cao, Y., Hu, H., Wang, L.: Memory enhanced global-local aggregation
  for video object detection. In: CVPR (2020)

\bibitem{cityscapes}
Cordts, M., Omran, M., Ramos, S., Rehfeld, T., Enzweiler, M., Benenson, R.,
  Franke, U., Roth, S., Schiele, B.: The cityscapes dataset for semantic urban
  scene understanding. In: CVPR (2016)

\bibitem{dai2021general}
Dai, X., Jiang, Z., Wu, Z., Bao, Y., Wang, Z., Liu, S., Zhou, E.: General
  instance distillation for object detection. In: CVPR (2021)

\bibitem{denil2013PredictingPI}
Denil, M., Shakibi, B., Dinh, L., Ranzato, M., de~Freitas, N.: Predicting
  parameters in deep learning. In: NeurIPS (2013)

\bibitem{distillationsurvey}
Gou, J., Yu, B., Maybank, S.J., Tao, D.: Knowledge distillation: A survey. IJCV
   (2021)

\bibitem{guo2020online}
Guo, Q., Wang, X., Wu, Y., Yu, Z., Liang, D., Hu, X., Luo, P.: Online knowledge
  distillation via collaborative learning. In: CVPR (2020)

\bibitem{quant2}
Gupta, S., Agrawal, A., Gopalakrishnan, K., Narayanan, P.: Deep learning with
  limited numerical precision. In: ICML (2015)

\bibitem{skipconv}
Habibian, A., Abati, D., Cohen, T.S., Bejnordi, B.E.: Skip-convolutions for
  efficient video processing. In: CVPR (2021)

\bibitem{comp2}
He, Y., Zhang, X., Sun, J.: Channel pruning for accelerating very deep neural
  networks. In: ICCV (2017)

\bibitem{hinton}
Hinton, G., Vinyals, O., Dean, J.: Distilling the knowledge in a neural
  network. arXiv preprint arXiv:1503.02531  (2015)

\bibitem{ddrnet}
Hong, Y., Pan, H., Sun, W., Jia, Y., et~al.: Deep dual-resolution networks for
  real-time and accurate semantic segmentation of road scenes. arXiv preprint
  arXiv:2101.06085  (2021)

\bibitem{hu2020tdnet}
Hu, P., Caba, F., Wang, O., Lin, Z., Sclaroff, S., Perazzi, F.: Temporally
  distributed networks for fast video semantic segmentation. CVPR  (2020)

\bibitem{fanet}
Hu, P., Perazzi, F., Heilbron, F.C., Wang, O., Lin, Z., Saenko, K., Sclaroff,
  S.: Real-time semantic segmentation with fast attention. ICRA  (2020)

\bibitem{quant3}
Jacob, B., Kligys, S., Chen, B., Zhu, M., Tang, M., Howard, A., Adam, H.,
  Kalenichenko, D.: Quantization and training of neural networks for efficient
  integer-arithmetic-only inference. In: CVPR (2018)

\bibitem{spatial_svd}
Jaderberg, M., Vedaldi, A., Zisserman, A.: Speeding up convolutional neural
  networks with low rank expansions. BMVC  (2014)

\bibitem{accel}
Jain, S., Wang, X., Gonzalez, J.E.: Accel: A corrective fusion network for
  efficient semantic segmentation on video. In: CVPR (2019)

\bibitem{jang2016categorical}
Jang, E., Gu, S., Poole, B.: Categorical reparameterization with
  gumbel-softmax. ICLR  (2017)

\bibitem{krish2018quant}
Krishnamoorthi, R.: Quantizing deep convolutional networks for efficient
  inference: {A} whitepaper. arXiv preprint arXiv:1806.08342  (2018)

\bibitem{zhu2018knowledge}
Lan, X., Zhu, X., Gong, S., et~al.: Knowledge distillation by on-the-fly native
  ensemble. NeurIPS  (2018)

\bibitem{lei2020blind}
Lei, C., Xing, Y., Chen, Q.: Blind video temporal consistency via deep video
  prior. NeurIPS  (2020)

\bibitem{li2017PruningFF}
Li, H., Kadav, A., Durdanovic, I., Samet, H., Graf, H.P.: Pruning filters for
  efficient convnets. arXiv preprint arXiv:1608.08710  (2017)

\bibitem{li2018low}
Li, Y., Shi, J., Lin, D.: Low-latency video semantic segmentation. In: CVPR
  (2018)

\bibitem{liu2018mobile}
Liu, M., Zhu, M.: Mobile video object detection with temporally-aware feature
  maps. In: CVPR (2018)

\bibitem{memory_guided}
Liu, M., Zhu, M., White, M., Li, Y., Kalenichenko, D.: Looking fast and slow:
  Memory-guided mobile video object detection. arXiv preprint arXiv:1903.10172
  (2019)

\bibitem{liu2019looking}
Liu, M., Zhu, M., White, M., Li, Y., Kalenichenko, D.: Looking fast and slow:
  Memory-guided mobile video object detection. arXiv preprint arXiv:1903.10172
  (2019)

\bibitem{liu2020efficient}
Liu, Y., Shen, C., Yu, C., Wang, J.: Efficient semantic video segmentation with
  per-frame inference. ECCV  (2020)

\bibitem{maddison2016concrete}
Maddison, C.J., Mnih, A., Teh, Y.W.: The concrete distribution: A continuous
  relaxation of discrete random variables. ICLR  (2017)

\bibitem{patchnet}
Mao, H., Zhu, S., Han, S., Dally, W.J.: Patchnet--short-range template matching
  for efficient video processing. arXiv preprint arXiv:2103.07371  (2021)

\bibitem{donna}
Moons, B., Noorzad, P., Skliar, A., Mariani, G., Mehta, D., Lott, C.,
  Blankevoort, T.: Distilling optimal neural networks: Rapid search in diverse
  spaces. In: ICCV (2021)

\bibitem{nagel2019dfq}
Nagel, M., van Baalen, M., Blankevoort, T., Welling, M.: Data-free quantization
  through weight equalization and bias correction. ICCV  (2019)

\bibitem{swiftnet}
Orsic, M., Kreso, I., Bevandic, P., Segvic, S.: In defense of pre-trained
  imagenet architectures for real-time semantic segmentation of road-driving
  images. In: CVPR (2019)

\bibitem{Rebol_2020_ACVRW}
Rebol, M., Knöbelreiter, P.: Frame-to-frame consistent semantic segmentation.
  In: Joint Austrian Computer Vision And Robotics Workshop (ACVRW) (2020)

\bibitem{fasterrcnn}
Ren, S., He, K., Girshick, R., Sun, J.: Faster r-cnn: Towards real-time object
  detection with region proposal networks. NeurIPS  (2015)

\bibitem{romera2017erfnet}
Romera, E., Alvarez, J.M., Bergasa, L.M., Arroyo, R.: Erfnet: Efficient
  residual factorized convnet for real-time semantic segmentation. IEEE
  Transactions on Intelligent Transportation Systems  (2017)

\bibitem{fitnets}
Romero, A., Ballas, N., Kahou, S.E., Chassang, A., Gatta, C., Bengio, Y.:
  Fitnets: Hints for thin deep nets. ICLR  (2015)

\bibitem{vid}
Russakovsky, O., Deng, J., Su, H., Krause, J., Satheesh, S., Ma, S., Huang, Z.,
  Karpathy, A., Khosla, A., Bernstein, M., Berg, A.C., Fei-Fei, L.: {ImageNet
  Large Scale Visual Recognition Challenge}. IJCV  (2015)

\bibitem{mobilenetv2}
Sandler, M., Howard, A., Zhu, M., Zhmoginov, A., Chen, L.C.: Mobilenetv2:
  Inverted residuals and linear bottlenecks. In: CVPR (2018)

\bibitem{shrivastava2016training}
Shrivastava, A., Gupta, A., Girshick, R.: Training region-based object
  detectors with online hard example mining. In: CVPR (2016)

\bibitem{sibechi2019exploiting}
Sibechi, R., Booij, O., Baka, N., Bloem, P.: Exploiting temporality for
  semi-supervised video segmentation. In: ICCV Workshops (2019)

\bibitem{tan2019efficientnet}
Tan, M., Le, Q.: Efficientnet: Rethinking model scaling for convolutional
  neural networks. In: ICML (2019)

\bibitem{efficientdet}
Tan, M., Pang, R., Le, Q.V.: Efficientdet: Scalable and efficient object
  detection. In: CVPR (2020)

\bibitem{tao2020hierarchical}
Tao, A., Sapra, K., Catanzaro, B.: Hierarchical multi-scale attention for
  semantic segmentation. arXiv preprint arXiv:2005.10821  (2020)

\bibitem{hrnet}
Wang, J., Sun, K., Cheng, T., Jiang, B., Deng, C., Zhao, Y., Liu, D., Mu, Y.,
  Tan, M., Wang, X., Liu, W., Xiao, B.: Deep high-resolution representation
  learning for visual recognition. TPAMI  (2019)

\bibitem{fgfi}
Wang, T., Yuan, L., Zhang, X., Feng, J.: Distilling object detectors with
  fine-grained feature imitation. In: CVPR (2019)

\bibitem{wang2019lednet}
Wang, Y., Zhou, Q., Liu, J., Xiong, J., Gao, G., Wu, X., Latecki, L.J.: Lednet:
  A lightweight encoder-decoder network for real-time semantic segmentation.
  In: ICIP (2019)

\bibitem{wu2021peer}
Wu, G., Gong, S.: Peer collaborative learning for online knowledge
  distillation. In: AAAI (2021)

\bibitem{bisenetv2}
Yu, C., Gao, C., Wang, J., Yu, G., Shen, C., Sang, N.: Bisenet v2: Bilateral
  network with guided aggregation for real-time semantic segmentation. IJCV
  (2021)

\bibitem{bisenetv1}
Yu, C., Wang, J., Peng, C., Gao, C., Yu, G., Sang, N.: Bisenet: Bilateral
  segmentation network for real-time semantic segmentation. In: ECCV (2018)

\bibitem{zhang2016acceleratingvd}
Zhang, X., Zou, J., He, K., Sun, J.: Accelerating very deep convolutional
  networks for classification and detection. TPAMI  (2016)

\bibitem{zhang2018deep}
Zhang, Y., Xiang, T., Hospedales, T.M., Lu, H.: Deep mutual learning. In: CVPR
  (2018)

\bibitem{icnet}
Zhao, H., Qi, X., Shen, X., Shi, J., Jia, J.: Icnet for real-time semantic
  segmentation on high-resolution images. In: ECCV (2018)

\bibitem{zhao2017pspnet}
Zhao, H., Shi, J., Qi, X., Wang, X., Jia, J.: Pyramid scene parsing network.
  In: CVPR (2017)

\bibitem{flow_guided}
Zhu, X., Dai, J., Zhu, X., Wei, Y., Yuan, L.: Towards high performance video
  object detection for mobiles. arXiv preprint arXiv:1804.05830  (2018)

\bibitem{fgfa}
Zhu, X., Wang, Y., Dai, J., Yuan, L., Wei, Y.: Flow-guided feature aggregation
  for video object detection. In: ICCV (2017)

\bibitem{zhu17dff}
Zhu, X., Xiong, Y., Dai, J., Yuan, L., Wei, Y.: Deep feature flow for video
  recognition. In: CVPR (2017)

\end{thebibliography}
